\documentclass{bmvc2k}

\usepackage{amssymb}
\usepackage{graphicx}
\usepackage{multirow}

\title{SILT: Self-supervised Lighting Transfer Using Implicit Image Decomposition}

\addauthor{Nikolina Kubiak}{n.kubiak@surrey.ac.uk}{1}
\addauthor{Armin Mustafa}{armin.mustafa@surrey.ac.uk}{1}
\addauthor{Graeme Phillipson}{graeme.phillipson@bbc.co.uk}{2}
\addauthor{Stephen Jolly}{stephen.jolly@bbc.co.uk}{2}
\addauthor{Simon Hadfield}{s.hadfield@surrey.ac.uk}{1}

\addinstitution{
 University of Surrey\\
 Centre for Vision, Speech and Signal Processing (CVSSP)\\
 Guildford, Surrey, UK
}
\addinstitution{
 BBC Research \& Development\\
 Salford, Greater Manchester, UK
}

\runninghead{Kubiak \etal}{SILT: Self-supervised Implicit Lighting Transfer}

\def\etal{\emph{et al}\bmvaOneDot}

\begin{document}

\maketitle
\vspace{-0.4cm}
\begin{abstract}
We present \textit{SILT}, a Self-supervised Implicit Lighting Transfer method. Unlike previous research on scene relighting, we do not seek to apply arbitrary new lighting configurations to a given scene. Instead, we wish to transfer the lighting style from a database of other scenes, to provide a uniform lighting style regardless of the input. The solution operates as a two-branch network that first aims to map input images of any arbitrary lighting style to a unified domain, with extra guidance achieved through implicit image decomposition. We then remap this unified input domain using a discriminator that is presented with the generated outputs and the style reference, \textit{i.e.}\ images of the desired illumination conditions. Our method is shown to outperform supervised relighting solutions across two different datasets without requiring lighting supervision. The code and pre-trained models can be found \underline{\href{https://github.com/n-kubiak/SILT}{here}}.
\end{abstract}
\vspace{-0.1cm}


\vspace{-0.3cm}
\section{Introduction}
\begin{figure}[t]
\begin{center}
\includegraphics[width=12cm]{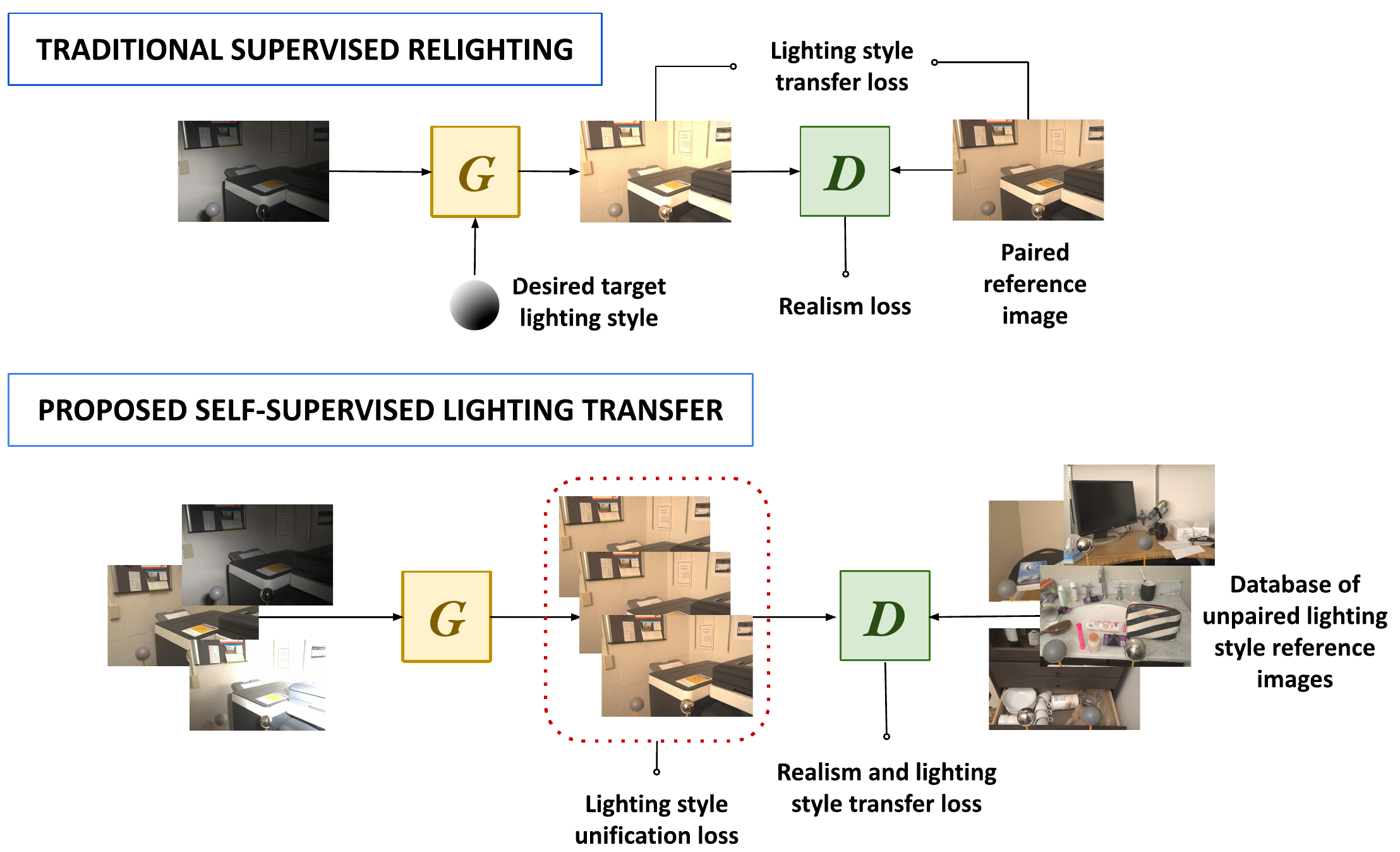}
\end{center}
\vspace{-0.5cm}
\caption{System overview: differences between relighting and lighting transfer.}
\label{basic_network}
\vspace{-0.3cm}
\end{figure}

\vspace{-0.2cm}
\label{sec:intro}
We propose the problem of lighting transfer where an input image under arbitrary illumination conditions is adapted to match the lighting of a style reference database. Previous approaches to a similar problem - arbitrary image relighting -  either make simplifying assumptions about the scene (\textit{e.g.}\ a single dominant object or a single light source) or require costly supervision where identical scenes must be captured under a large number of known lighting conditions. In contrast, our lighting transfer approach, \textit{SILT}, is entirely self-supervised. SILT requires only a training dataset with multiple examples of the same scene and a style reference database. It is not necessary to know the ground truth lighting conditions for the training data, nor is it necessary for the target illumination to be present within the training dataset. This distinction between image relighting and lighting transfer is demonstrated in Fig.\ \ref{basic_network}. \looseness -1

The proposed SILT method consists of a two-branch network. During training, the generator sees a number of input images of the same scene under different unknown illuminations. The model attempts to enforce similarity between the illumination conditions of the outputs while preserving the contents of the scene. The generated data is then presented to a discriminator that tries to distinguish the outputs from the examples of target illumination (\textit{i.e.}\ the reference style images), thus enforcing the desired lighting conditions. At inference, SILT takes in a single input image and can adapt to previously unseen locations. 

We motivate our work by the fact that in complex, real-life scenarios there is rarely one \textit{correct} way to describe the desired lighting conditions. Instead of trying to quantify a particular style (\textit{e.g.}\ a TED talk), we can show the model examples of such an event. We do not need the re-styled materials to look like a specific, \textit{best} talk - we just want the model to learn what such talks look like and copy the style. The same idea can be applied to other domains such as casual live music performances, filters for social media or domain adaptation for multi-view reconstruction. In summary, the contributions of our paper are threefold: \looseness -1
\begin{itemize}
    \vspace{-0.15cm}
    \item We present the first formulation of the lighting transfer problem, with associated baselines and evaluation protocol.
    \vspace{-0.15cm}
    \item We demonstrate a flexible self-supervised approach to lighting transfer, performing on par with supervised relighting models.
    \vspace{-0.15cm}
    \item We show that jointly training a self-supervised image decomposition step allows us to better capture the complexities of real scenes, compared to a pre-trained model.
    \vspace{-0.3cm}
\end{itemize}

\vspace{-0.2cm}
\section{Literature review}
As lighting transfer is a new field of inquiry, we will instead focus this discussion on the most relevant works from the field of image relighting. 
    \vspace{-0.2cm}
\newline\newline \textbf{\textcolor{bmv@sectioncolor}{General-purpose relighting. }}Last year the VIDIT dataset \cite{vidit} was created and used for relighting challenges \cite{helou_aim_2020, helou_ntire_2021}. It features indoor and outdoor scenes lit from 8 directions and captured under 5 temperature settings. 
The applications of VIDIT include \cite{wang_deep_2020} where Wang \etal\ tackle the task of relighting in stages. They simultaneously teach two network bran-ches to remove the effects of source illumination and to apply shadows according to target illumination, and then combine them to render the final result. In \cite{das_msr-net_2020} Das \etal\ stack two U-Nets to achieve good quality reconstructions while keeping the computational cost very low. They expand on this idea in \cite{das_dsrn_2021} and use 2 images captured from opposite directions as input, which allows for better reconstruction of dark regions. Gafton and Maraz \cite{gafton2020} expand the pix2pixHD \cite{pix2pixHD} architecture by adding light direction estimation layers, and train 8 network branches that can relight any image to 1 of 8 illumination directions covered by VIDIT. Puthussery \etal\ \cite{puthussery_wdrn_2020} perform relighting using wavelet decomposition and propose a new loss responsible for accurate shadow recasting. Their \textit{gray loss} is calculated on blurred input and ground truth images, stripped from texture information to focus on illumination. As shown, in just a year VIDIT has found a number of applications. The dataset, however, is synthetic and unnaturally dark, making it difficult to apply models trained on it to real-life data. \looseness -1
 
Very few papers operate on real-life data captured in the wild. Zhou \etal\ \cite{zhou_relight_2015} perform image decomposition and then relight scenes by swapping the shading maps of two images. Nevertheless, the approach is limited to relighting within a fixed location, and the lighting style captured by the low-level shading map cannot be applied to another scene. Murmann \etal \cite{murmann19} created a multi-illumination indoor dataset, published alongside an encoder-decoder relighting model. 
\vspace{-0.15cm}
\newline\newline \textbf{\textcolor{bmv@sectioncolor}{Reducing supervision requirements. }}All of the previously described solutions operate in a supervised manner. Yet, such an approach makes training difficult as pairs of images of the same scene, captured under source and target illumination conditions, are rarely readily available outside of simulation. Therefore, a number of authors have recently begun to explore domain-specific mechanisms to reduce the supervision requirements of relighting.

Liu \etal\ \cite{liu_relighting_2020} propose a Siamese autoencoder network for portrait relighting that decomposes the images into illumination and content embeddings. The process is supervised by augmenting the available data in a way that preserves its content while changing the light direction, thus creating training pairs. The relighting task is constrained via a simplified spherical harmonic lighting model. This, however, limits the relighting to light direction and intensity changes, and neglects the impact of colour on the white-balance of the outputs. \looseness -1

Liu \etal \cite{liu_learning_2020} and Yu \etal\ \cite{yu_self-supervised_2020} proposed self-supervised image-based outdoor relighting models that first decompose the scene into constant and changing factors, and then synthesise new images by recombining the desired illumination conditions with the illumination-independent elements. Yu \etal\ show that during decomposition, phenomena such as shadows or specularities tend to be wrongly preserved in reflectance or normal maps, skewing the final renderings. To address this, they expand on their previous work \cite{yu_and_smith_2019} and add a channel that aims to capture this type of residual information. Inspired by \cite{li_learning_2018}, Liu \etal\ \cite{liu_learning_2020} learn from stacks of timelapse panoramas of the same location. They capture time-varying information using high-level features, allowing the system to mix and match illumination settings between locations. However, the method requires the target lighting conditions to be present in a panorama stack, which limits the model's flexibility.

\section{Methodology}
\subsection{Image relighting posed as style transfer}
Our model is an LSGAN \cite{lsgan} based on the pix2pix \cite{pix2pixHD} U-Net \cite{ronneberger_2015_unet} architecture. The multi-scale discriminator operates on patches and distinguishes between real and fake samples at different spatial scales. Our generator's architecture consists of 3 downsampling steps, 9 residual blocks and 3 upsampling steps, and does not require the LocalEnhancer add-on proposed in the pix2pix paper. To achieve self-supervision, the model structure is duplicated to create two branches; we refer to them as branches A and B.

The training of our generator is guided by a sum of three types of losses. Firstly, we use a generator GAN loss $\mathcal{L}_{g}$ which measures the model's ability to fool the discriminator. It has the opposite goal and acts as an adversary to the discriminator loss $\mathcal{L}_{d}$ which ensures that the real and generated samples are classified as such. In our case, this distinction covers both image realism and the illumination style transfer accuracy. The losses can be expressed as
\begin{equation}
   \mathcal{L}_{g} = \lVert D\left(G\left(I\right)\right)- 1 \rVert_2 \quad\mathrm{and}\quad  \mathcal{L}_{d} = \lVert D\left(\tilde{I}\right) - 1 \rVert_2 + \lVert D\left(G\left(I\right)\right)\rVert_2,
    \label{gan_loss}
\end{equation}
where $\tilde{I}$ and $I$ correspond to reference and input samples, and $D$ and $G$ mark the discriminator and generator models. 

Now, to enforce the characteristics of the target illumination, we need to first encourage the generator to translate all images, regardless of their initial lighting conditions, to a unified domain. To this end, we enforce similarity between the outputs $\hat{I}$ of both branches using output similarity loss $\mathcal{L}_{os}$ defined as
\begin{equation}
    \mathcal{L}_{os} = \lVert \hat{I}_A - \hat{I}_B \rVert_1 + \lVert \nabla{\hat{I}_A} - \nabla{\hat{I
    }_B} \rVert_1  = \lVert G\left(I_A\right) - G\left(I_B\right) \rVert_1 + \lVert \nabla{G\left(I_A\right)} - \nabla{G\left(I_B\right)} \rVert_1.
    \label{os_loss}
\end{equation}
In the equation, $\nabla{\hat{I}}$ symbolises spatial gradients calculated over the generated images. We choose these two types of losses to ensure the correct matching of colour and brightness as well as edge information.

Finally, to prevent mode collapse, we need to make the output have the same visual content as the original image. This is controlled using the content preservation loss $\mathcal{L}_{cp}$, calculated between the input and output for each branch. This error can be described as
\begin{equation}
    \mathcal{L}_{cp} = \mathcal{VGG}\left(I,\hat{I}\right) = \mathcal{VGG}\left(I, G\left(I\right)\right),
    \label{cp_loss}
\end{equation}
where $\mathcal{VGG}$ represents the perceptual loss proposed in \cite{chen_2017_perc}. This is a weighted L1 loss calculated between the features extracted from input and output images using specific layers of a pre-trained VGG-19 network. 

To summarise, the described model performs lighting style transfer, controlling the content and style of the output image. The former is preserved using a perceptual loss $\mathcal{L}_{cp}$. For the latter, we propose a new approach and split the task between the generator and discriminator. The former uses $\mathcal{L}_{os}$ to ensure a unified style, and the latter controls the characteristics of said style by comparing generated data with style reference examples.


\subsection{Decomposition-guided relighting}
In addition to the simple style transfer approach, we propose to enhance the generator with a jointly learnt self-supervised image decomposition mechanism. The decomposition process, inspired by the Retinex theory \cite{land_retinex_1977}, assumes that an image $I$ can be expressed as a Hadamard product of its reflectance $R$ and shading $S$, \textit{i.e.}\ $I = R \odot S$. These correspond to the illumination-invariant vs the lighting-dependent factors, respectively. 

There are many general purpose solutions that use inverse rendering techniques to achieve this kind of decomposition \cite{nestmeyer_reflectance_2017, li_learning_2018, li_cgintrinsics_2018, baslamisli_physics-based_2020, barron_shape_2020, sengupta_neural_2019, wei_deep_2018}. Of particular interest to us is the state-of-the-art decomposition model by Liu \etal\ \cite{liu_unsupervised_2020}, combining indoor scenes and unsupervised learning. The authors propose a scheme in which the style of reflectance and shading images is learnt based on unpaired samples. At the same time, their generator $G_{\lambda}$ learns to decompose images into underlying factors that match their corresponding learnt distributions, \textit{i.e.}\ $\{R,S\} = G_{\lambda}(I)$. In our use case, we do not require a set of reference reflectance and shading maps. We simply apply self-supervision asking the network to extract two features from the image that, when multiplied, produce an accurate reconstruction of the original input. As such they should serve roles similar to reflectance and shading yet this is never defined explicitly. Nevertheless, in the interest of clarity, we will use the reflectance-shading naming convention throughout the paper. 
 \begin{figure}[t]
\begin{center}
\includegraphics[width=12cm]{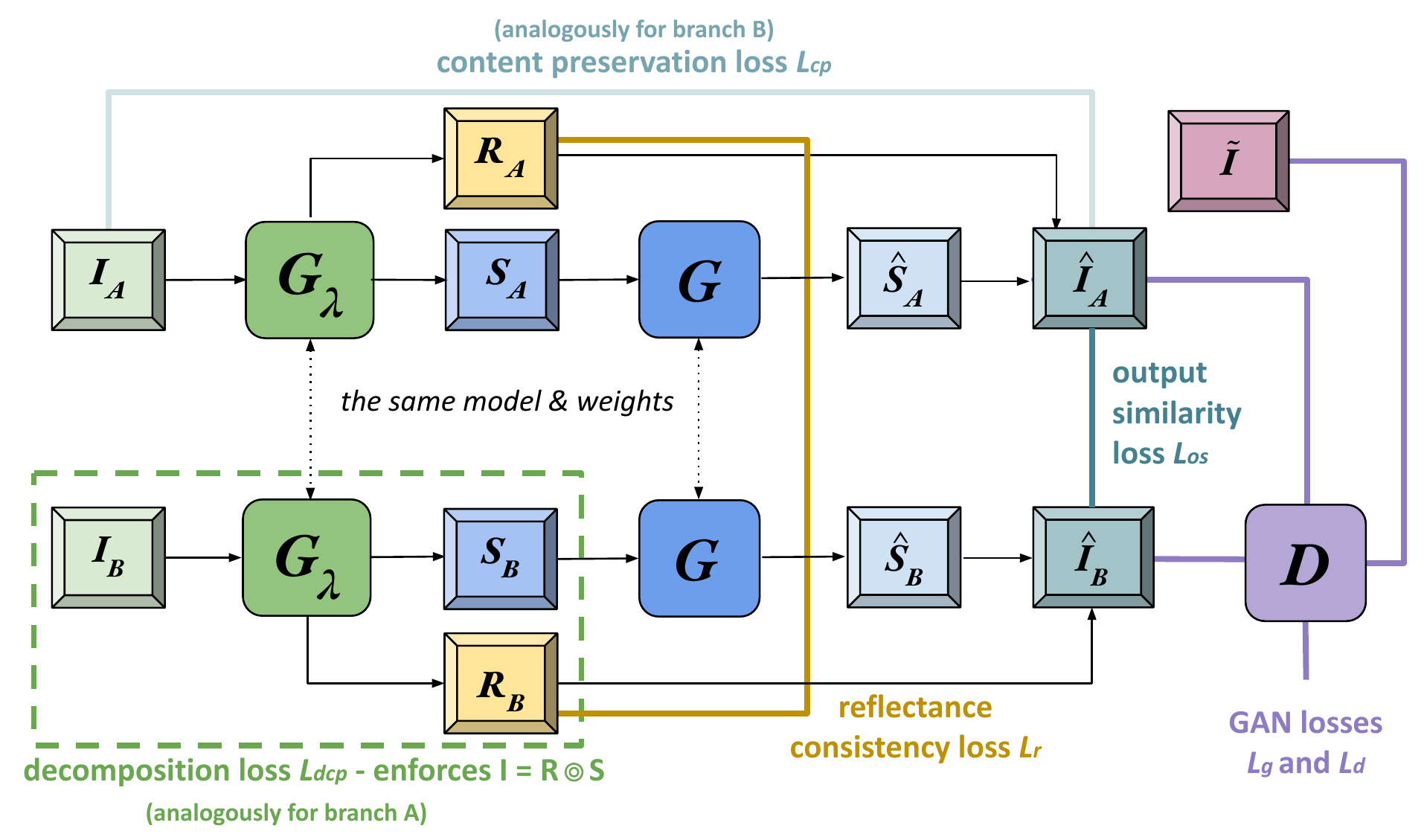}
\end{center}
\caption{All parts and losses of SILT, the implicit decomposition model.}
\vspace{-0.3cm}
\label{diagram}
\vspace{-0.1cm}
\end{figure}

During training, our network must learn to decompose images and to optimise this procedure to support the downstream lighting transfer task. We guide the process by minimising the residual between the input image and the product of its generated reflectance and shading maps. This is done using the decomposition loss $\mathcal{L}_{dcp}$, expressed as
\begin{equation}
    \mathcal{L}_{dcp} = \lVert I - R \odot S \rVert_1.
    \label{decomp_loss}
\end{equation}

Since we only want to perform the style transfer on the shading component $S$, we feed its corresponding map into the generator $G$ instead of using the original image. The reflectance maps should remain unchanged so these are passed via a skip connection and multiplied with the newly generated shading image $\hat{S}$ to get the output image, as shown in Fig.\ \ref{diagram}. The output image formulation can now be described as $\hat{I} = R \odot \hat{S} = R \odot G\left(S\right)$.


Additionally, we want to enforce consistency between reflectance maps generated for both branches as they should be identical regardless of the original illumination conditions. We control this using the reflectance loss $\mathcal{L}_{r}$ expressed as
\begin{equation}
    \mathcal{L}_{r} = \lVert R_A - R_B \rVert_2, 
    \label{rs_loss}
\end{equation}
where $R_A$ and $R_B$ are the reflectances obtained during decomposition for each input pair. 

\section{Experiments}

\subsection{Ablation study}

During these experiments all models were trained on the Multi-Illumination dataset \cite{murmann19}. Its data is split into train and test sets, with 985 and 30 scenes respectively. Each scene is photographed under 25 fixed lighting directions located in the upper hemisphere relative to the camera. For this paper we picked nine illumination conditions, consistent across all scenes of the dataset, and chose one unpaired image setting to act as target lighting domain. During training, we paired the nine images up and each unique pair was presented to the generator once per epoch. At test time, the models were fed single images of previously unseen scenes, captured under the same nine illumination conditions; no additional reference data was needed. For our experiments the images were resized to 768 $\times$ 512 pixels. 

Our method was trained for approx.\ 180k iterations and used the same default training hyperparameters as the pix2pix GAN; this holds true for all discussed model versions. We report additional information regarding the model's complexity in Appendix C. In the following sections we measure the performance of the tested models using SSIM and PSNR metrics (higher is better) and report the perceptual (VGG) loss scores (lower is better). For results reported in tables, the top score is shown in bold and the second best is underlined.
\newline\newline \textbf{\textcolor{bmv@sectioncolor}{Decomposition.}}
First, we explore the benefits of the jointly-learnt self-supervised decomposition approach. We compare our basic model with no decomposition against our network using the fixed pre-trained decomposition approach of Liu \etal\ \cite{liu_unsupervised_2020} and our full model with a jointly-learnt decomposition step based on Liu \etal's architecture. For the pre-trained model, we used the generated reflectances as-is but also tried averaging them across the scene to support reflectance consistency. Below we show results for the latter, more successful approach. For the $R$ and $S$ maps generated during the jointly-learnt decomposition step, please see Appendix A.

\begin{figure}[h]
    \begin{center}
    \begin{tabular}{c @{\hspace{0.05cm}} c @{\hspace{0.05cm}}c@{\hspace{0.05cm}}c@{\hspace{0.05cm}}c}
    \footnotesize Input & \footnotesize None &\footnotesize Liu \etal &\footnotesize Learnt (SILT)&\footnotesize Ground truth\\
    \includegraphics[width=2.45cm]{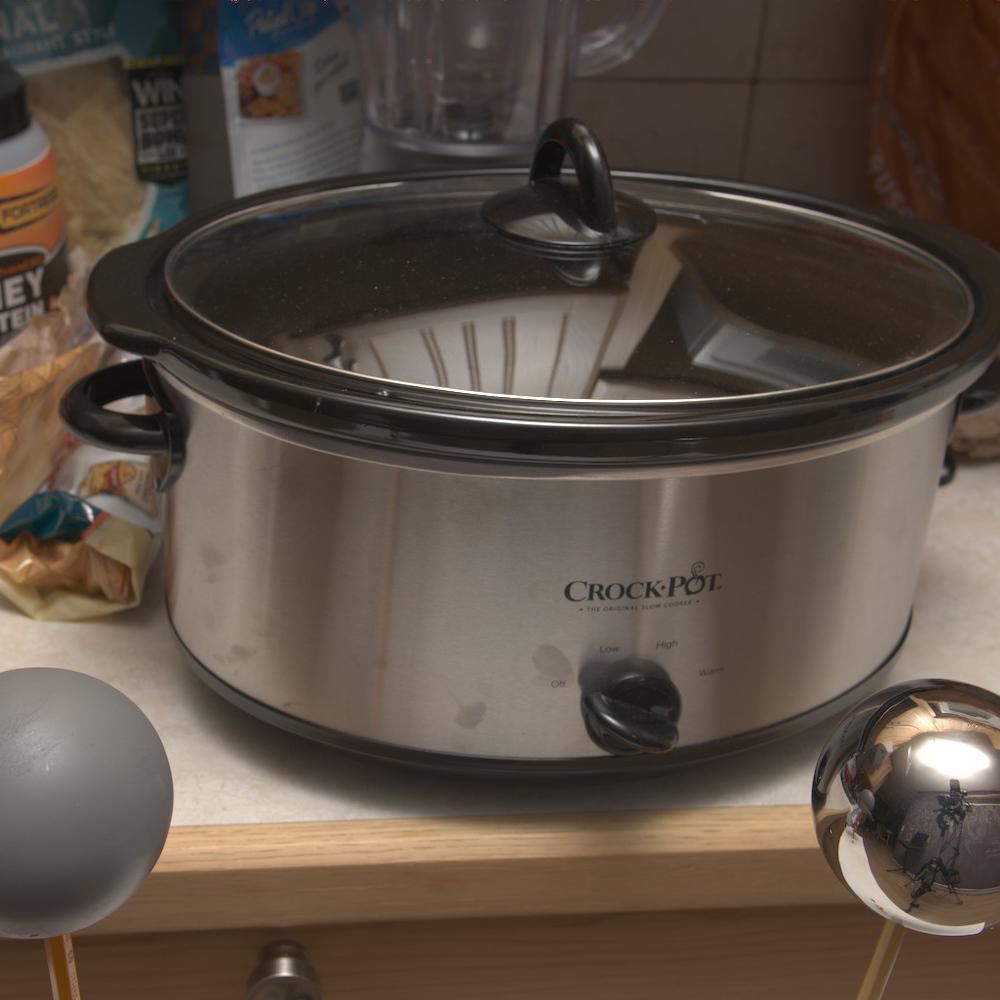}&
    \includegraphics[width=2.45cm]{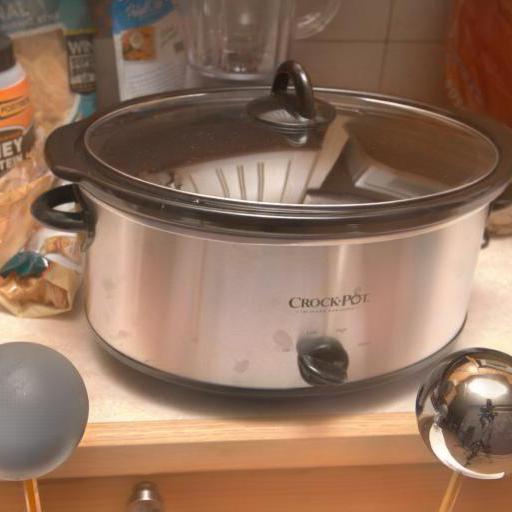}&
    \includegraphics[width=2.45cm]{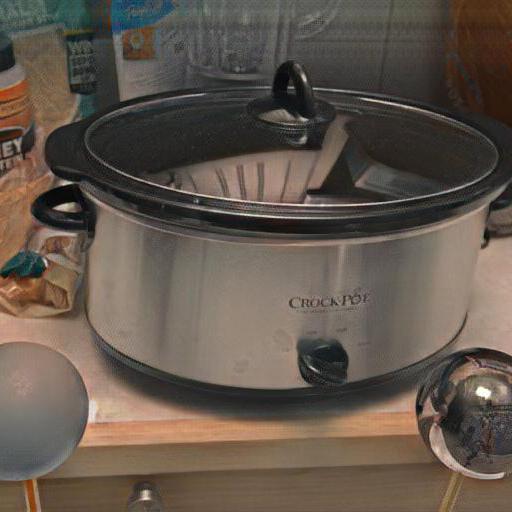}&
    \includegraphics[width=2.45cm]{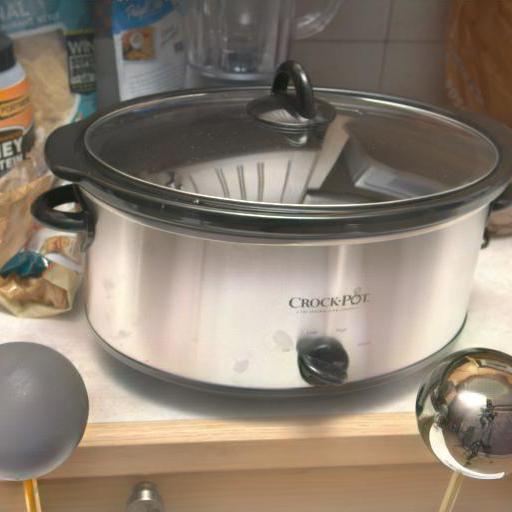}&
    \includegraphics[width=2.45cm]{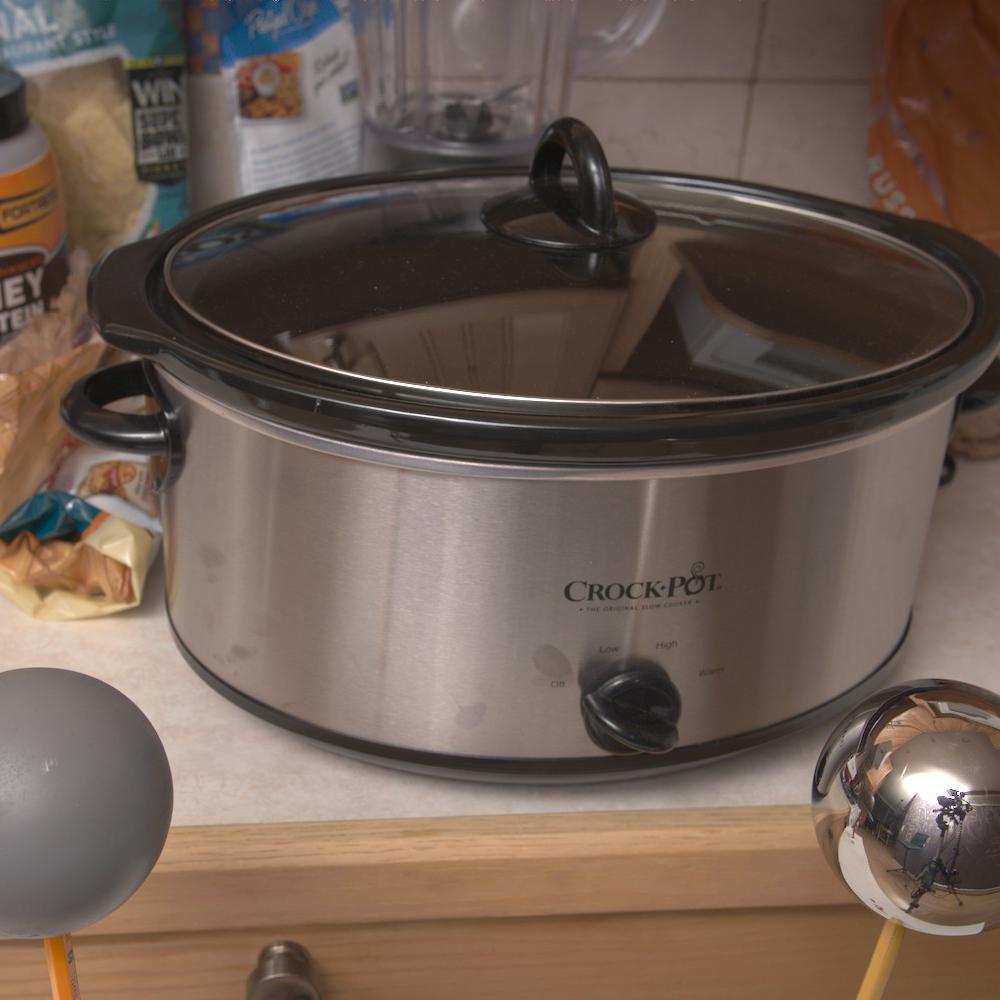}\\
    \includegraphics[width=2.45cm]{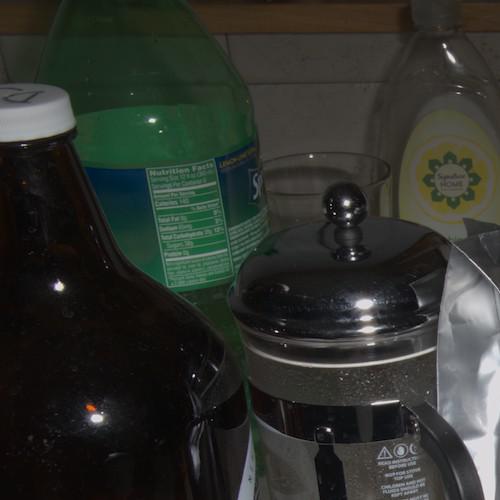}&
    \includegraphics[width=2.45cm]{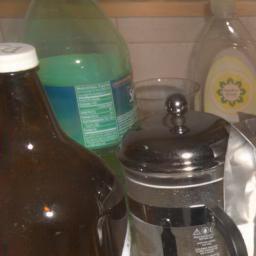}&
    \includegraphics[width=2.45cm]{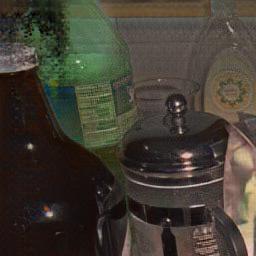}&
    \includegraphics[width=2.45cm]{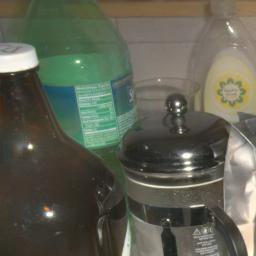}&
    \includegraphics[width=2.45cm]{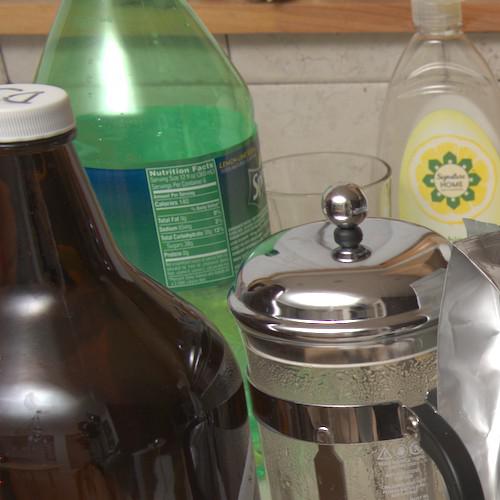}
    \end{tabular}
    \end{center}
    \caption{Visual results across different decomposition settings.}
    \label{mit_visual}
    \vspace{-0.5cm}
    \end{figure}

The outcomes of the study are measured quantitatively and qualitatively, with results in Table \ref{ablation_architecture} and Fig.\ \ref{mit_visual}. 
The basic version and the full implicit decomposition versions of our self-supervised model achieve similar performance across all metrics. The generated images, however, show that the latter network, SILT, provides a closer colour match and makes a better attempt at specularity reconstruction (bottom row). Even though the decomposition step of Liu \etal\ was pretrained on indoor scenes, it cannot produce results on par with our implicit decomposition model. Its generated images are too dark and contain more prominent artefacts. 
    \begin{table}[h]
    \begin{center}
    \begin{tabular}{ |l|c|c|c|  }
    \hline
    Decomposition &  SSIM $\uparrow$ & PSNR $\uparrow$ & VGG $\downarrow$ \\
    \hline \hline
    None &  \underline{0.800}&\underline{17.995}&\underline{0.323} \\
    Liu \etal\ \cite{liu_unsupervised_2020} & 0.689 & 15.874 &0.423 \\
    Learnt & \textbf{0.811}&\textbf{18.667}&\textbf{0.320}  \\
    \hline
    \end{tabular}
    \end{center} 
    \caption{Decomposition architecture study}
    \label{ablation_architecture}
    \end{table}
\newline \textbf{\textcolor{bmv@sectioncolor}{Losses.}} 
Next, we review the losses used by our best self-supervised solution, \textit{i.e.}\ the implicit decomposition model. We start with just the GAN losses - $\mathcal{L}_{g}$ and $\mathcal{L}_{d}$ - and the decomposition loss $\mathcal{L}_{dcp}$ as these are responsible for the core functionality of our network. Then, we gradually add the other losses. The changes linked to each component are presented in Table \ref{ablation_losses}. \looseness -1

\newpage The results show that the most meaningful improvements in the SSIM and VGG metrics stem from the content preservation loss $\mathcal{L}_{cp}$. Interestingly, using just the output similarity loss $\mathcal{L}_{os}$ leads to a performance decrease. However, when this loss is linked with $\mathcal{L}_{cp}$ (line \#4), the performance across all metrics improves visibly. 
This synergy suggests that simply trying to equate two outputs without constraining their content creates an ill-posed task and more guidance is needed to generate meaningful renders. 
Finally, we can observe a slight increase in SSIM and larger improvements in PSNR when $\mathcal{L}_{r}$ is added. This proves that enforcing reflectance consistency between the branches leads to better $R$ and $S$ decomposition and, consequently, more accurate lighting style transfer achieved by the generator $G$. In the future experiments we will use the model using all losses, corresponding to line \#5 in the table. \looseness -1
    \begin{table}[h]
    \begin{center}
    \begin{tabular}{ |c|c|c|c|c|c|c|c|  }
    \hline
    \# & core & $\mathcal{L}_{os}$ & $\mathcal{L}_{cp}$ & $\mathcal{L}_{r}$ &  SSIM $\uparrow$ & PSNR $\uparrow$ & VGG $\downarrow$  \\
    \hline \hline
    1& \checkmark & & & & 0.730	&17.075	&0.428 \\
    2&\checkmark & \checkmark & & &  0.682 &17.142 & 0.495 \\
    3&\checkmark & &\checkmark & &  0.798&	17.575&	\underline{0.319} \\
    4&\checkmark &\checkmark & \checkmark&   &\underline{0.805} &\underline{17.712} & \textbf{0.317}\\
    5&\checkmark & \checkmark& \checkmark& \checkmark& \textbf{0.811}&\textbf{18.667}&0.320  \\
    \hline
    \end{tabular}
    \end{center}
    \caption{Loss ablation study}
    \label{ablation_losses}
    \vspace{-0.5cm}
    \end{table}
\subsection{State-of-the-art comparisons on the \textit{Multi-Illumination} dataset}
Next, we compare the performance of the best version of our self-supervised SILT model with the state-of-the-art supervised method by Murmann \etal\ \cite{murmann19} and the pix2pix baseline \cite{pix2pixHD}. Murmann \etal's solution originally focused on relighting from a `base' image to any of the desired lighting directions; we refer to this as \textit{one2any} translation. To adapt their approach to our target application, we retrained it in \textit{any2one} fashion, \textit{i.e.}\ so that it can convert any lighting condition seen during training to the reference illumination. The same approach is followed for pix2pix. We train the baseline with the default model settings and, since we do not train with HD data, only the global generator is used. It is worth noting that both of the comparisons are trained in a supervised manner. They still use the same 9 input light directions as our model, but these are paired with a target image of the same scene that our generator never sees.

\begin{figure}[t]
    \vspace{-0.2cm}
    \begin{center}
    \begin{tabular}{c @{\hspace{0.05cm}} c @{\hspace{0.05cm}}c@{\hspace{0.05cm}}c@{\hspace{0.05cm}}c}
    \vspace{-0.1cm}
    & \tiny SUPERVISED &\tiny SUPERVISED &\tiny SELF-SUPERVISED & \\
    \footnotesize Input & \footnotesize Murmann \etal &\footnotesize pix2pix &\footnotesize SILT (ours) & \footnotesize Ground truth\\
    \includegraphics[width=2.45cm]{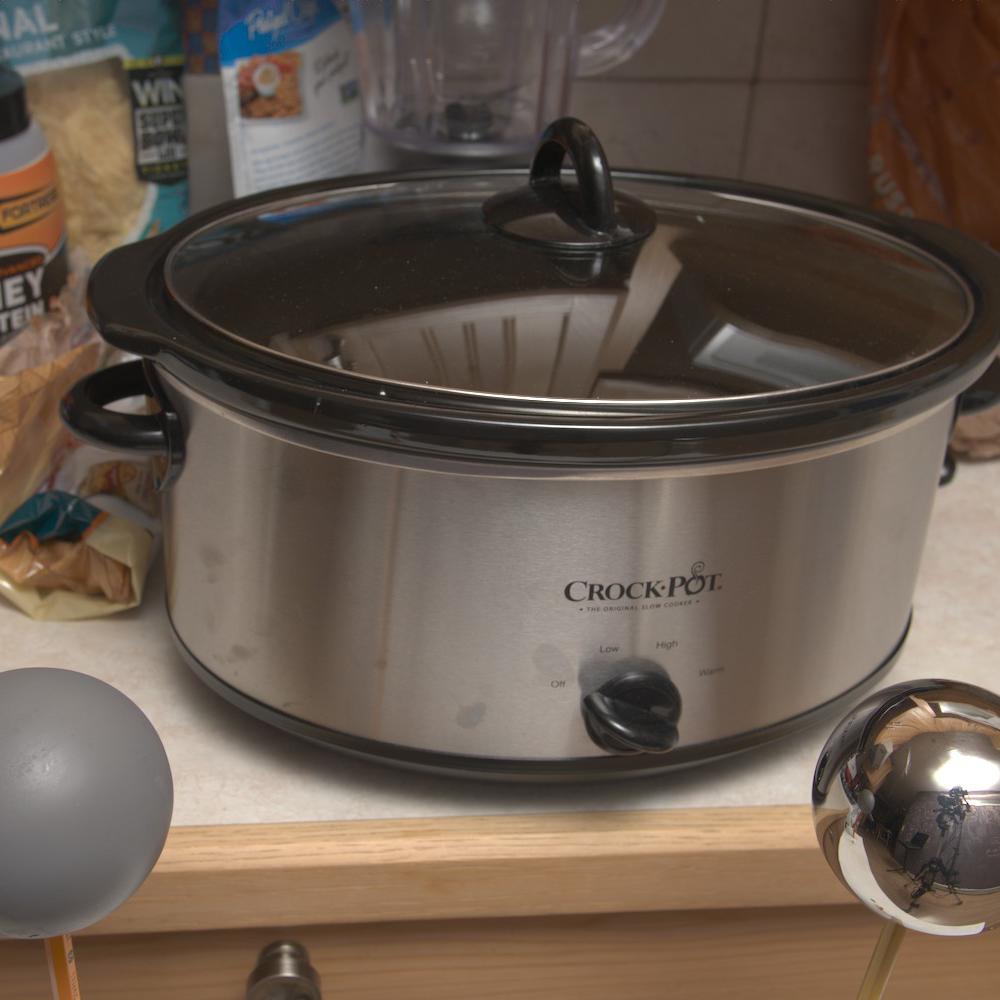}&
    \includegraphics[width=2.45cm]{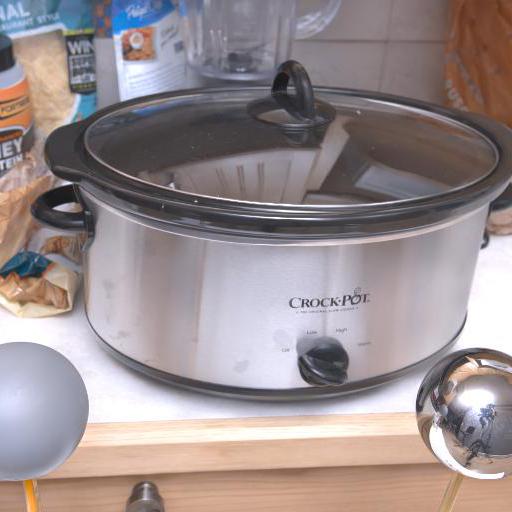}&
    \includegraphics[width=2.45cm]{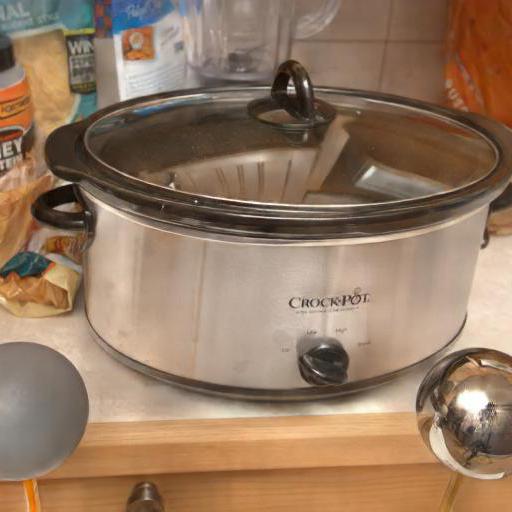}&
    \includegraphics[width=2.45cm]{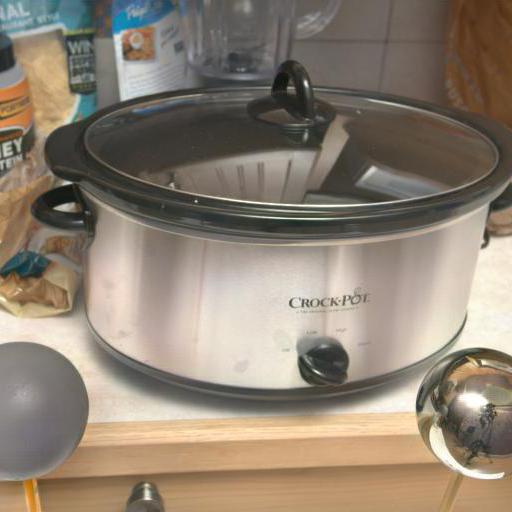}&
    \includegraphics[width=2.45cm]{mit/kitchen14_ref.jpg}\\
    \includegraphics[width=2.45cm]{mit/kitchen6_in2.jpg}&
    \includegraphics[width=2.45cm]{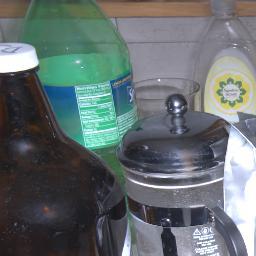}&
    \includegraphics[width=2.45cm]{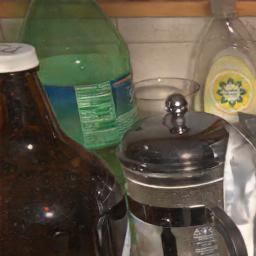}&
    \includegraphics[width=2.45cm]{mit/kitchen6_out2_SILT.jpg}&
    \includegraphics[width=2.45cm]{mit/kitchen6_ref.jpg}
    \end{tabular}
    \end{center}
    \vspace{-0.2cm}
    \caption{Visual results on the Multi-Illumination dataset.}
    \label{mit_comparison}
    \vspace{-0.5cm}
    \end{figure}

\newpage The results of the experiments are shown in Table \ref{mit} and Fig.\ \ref{mit_comparison}. SILT comes first across all three performance metrics. Our relit images have the best colour balance and compare favourably against Murmann \etal's colder tones and pix2pix's warm tint (top row). 
When it comes to specularities (bottom row), the model of Murmann \etal\ appears to smudge them. The pix2pix baseline does not make any changes. Our solution, on the other hand, shifts the input reflection to the left, as in the reference image. Since specularities are particularly hard to re-render, we believe this is a fair reconstruction attempt.

    \begin{table}[h]
    \vspace{-0.1cm}
    \begin{center}
    \begin{tabular}{ |l|c|c|c|  }
    \hline
    Model &  SSIM $\uparrow$ & PSNR $\uparrow$ & VGG $\downarrow$\\
    \hline \hline
    Murmann \etal\ \cite{murmann19} &   0.763 & 13.951&0.381\\
    pix2pix &  \underline{0.799} & \underline{18.534} & \underline{0.333}\\
    SILT (ours)&\textbf{0.811}&\textbf{18.667}&\textbf{0.320}   \\
    \hline
    \end{tabular}
    \end{center}
    \caption{Results on the Multi-Illumination dataset}
    \label{mit}
    \vspace{-0.5cm}
    \end{table}
\subsection{State-of-the-art comparisons on the \textit{VIDIT} dataset}
\label{VIDIT_section}
To make comparisons against VIDIT-based models, we retrain SILT on the synthetic VIDIT dataset, and compare our work against recent supervised methods. For this purpose, we choose two supervised solutions trained in \textit{any2one} fashion that share quantitative results in their corresponding papers and run evaluations on the publicly available sets of VIDIT. Additionally, we train pix2pix on VIDIT and use it as a baseline.

The chosen methods differ slightly in their approach to the task. Gafton and Maraz \cite{gafton2020} split the VIDIT training set 90:10 into train:test sets, and translate the images within one temperature setting (4500K). They separately measure the ability of their model to relight to all of the 8 directions. For our comparisons in this paper we pick one of them - East (E). Wang \etal\ \cite{wang_deep_2020} train their model to convert all input directions and temperatures of the training set to 4500K-E, and then during inference convert from 6500K-N to 4500K-E (as in the AIM2020 challenge \cite{helou_aim_2020}). In the interest of fairness, we train two versions of our self-supervised model and the baseline to follow the same rules as mentioned above.

\begin{figure}[b!]
    \vspace{-0.4cm}
    \begin{center}
    \begin{tabular}{c @{\hspace{0.05cm}} c @{\hspace{0.05cm}}c@{\hspace{0.048cm}}c@{\hspace{0.05cm}}c}
    \vspace{-0.1cm}
    & \tiny SUPERVISED &\tiny SUPERVISED &\tiny SELF-SUPERVISED & \\
    \footnotesize Input & \footnotesize Gafton \& Maraz &\footnotesize pix2pix &\footnotesize SILT (ours) &\footnotesize Ground truth\\
    \includegraphics[width=2.45cm]{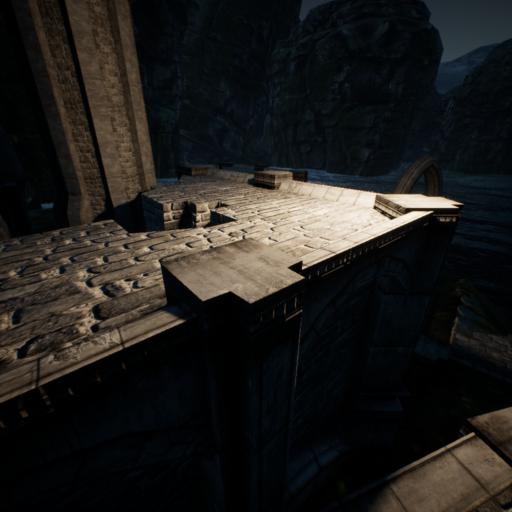}&
    \includegraphics[width=2.45cm]{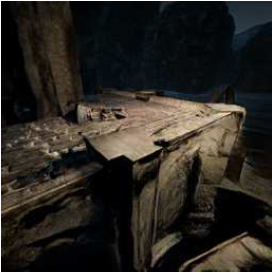}&
    \includegraphics[width=2.45cm]{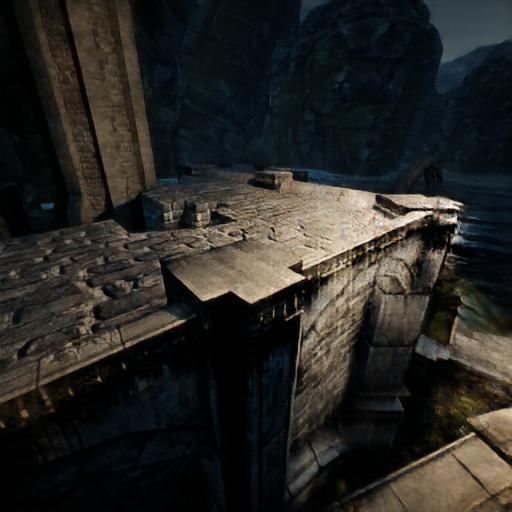}&
    \includegraphics[width=2.45cm]{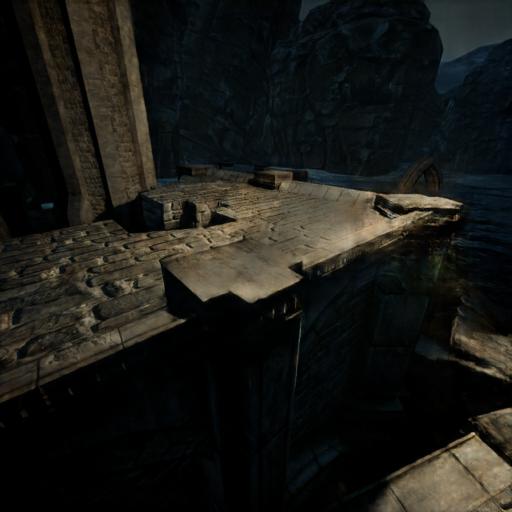}&
    \includegraphics[width=2.45cm]{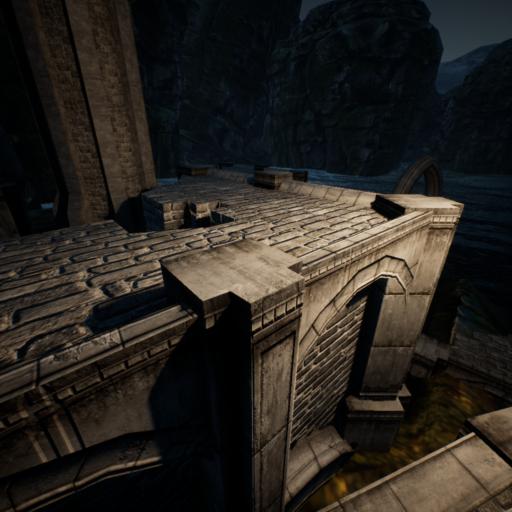}\\
    \vspace{-0.1cm}
    & \tiny SUPERVISED &\tiny SUPERVISED &\tiny SELF-SUPERVISED & \\
    \footnotesize Input & \footnotesize Wang \etal &\footnotesize pix2pix &\footnotesize SILT (ours) & \footnotesize Ground truth\\
    \includegraphics[width=2.45cm]{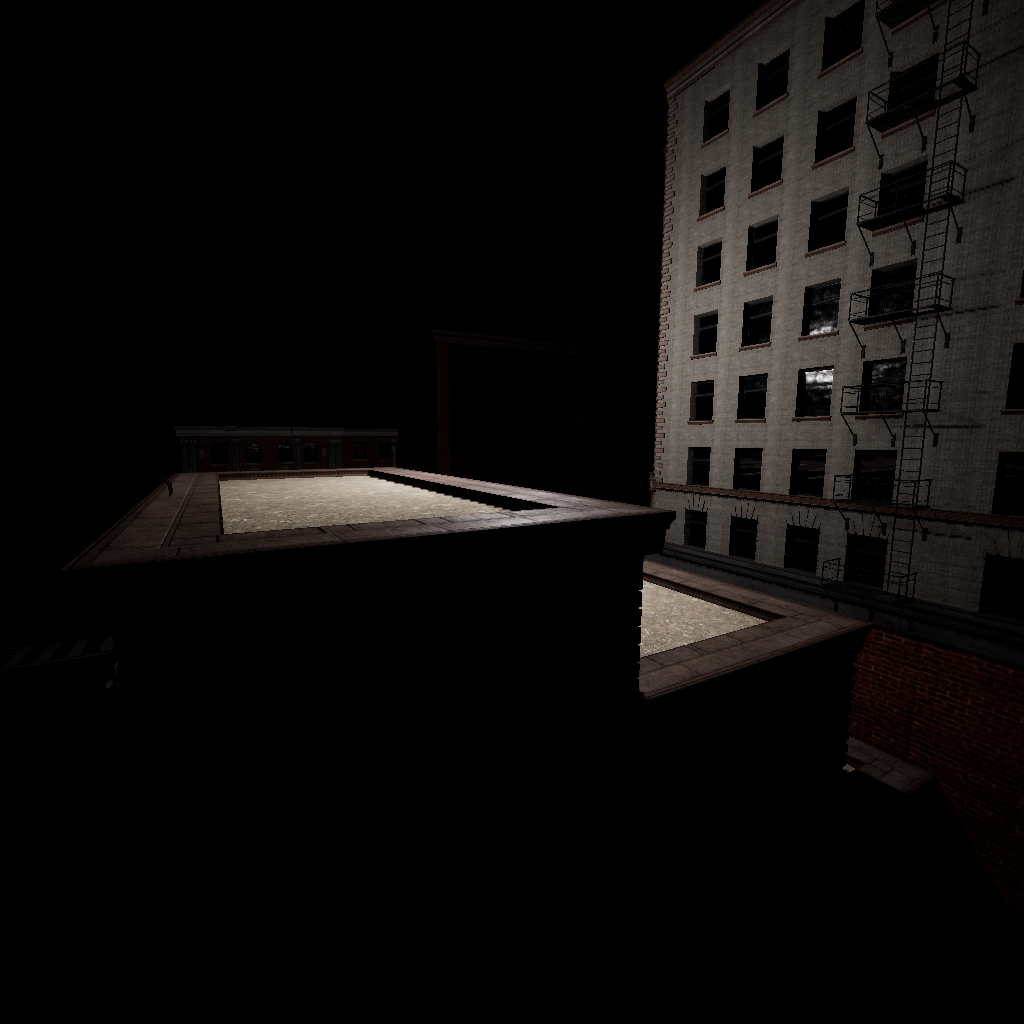}&
    \includegraphics[width=2.45cm]{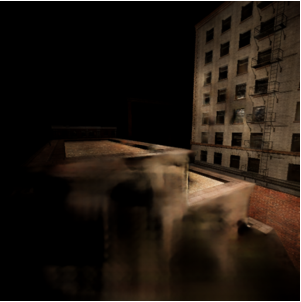}&
    \includegraphics[width=2.45cm]{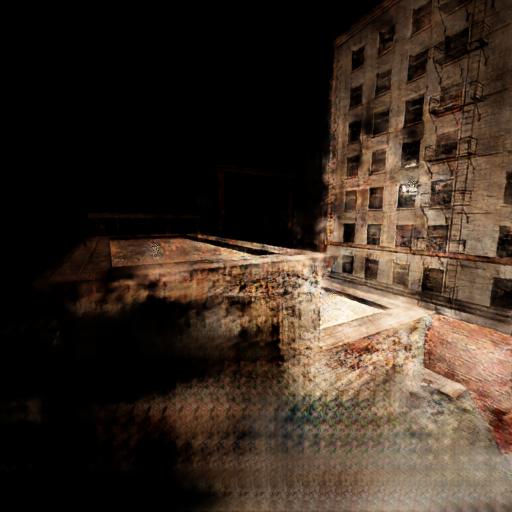}&
    \includegraphics[width=2.45cm]{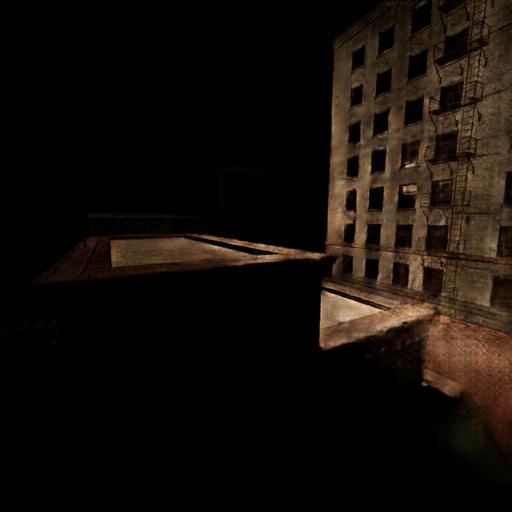}&
    \includegraphics[width=2.45cm]{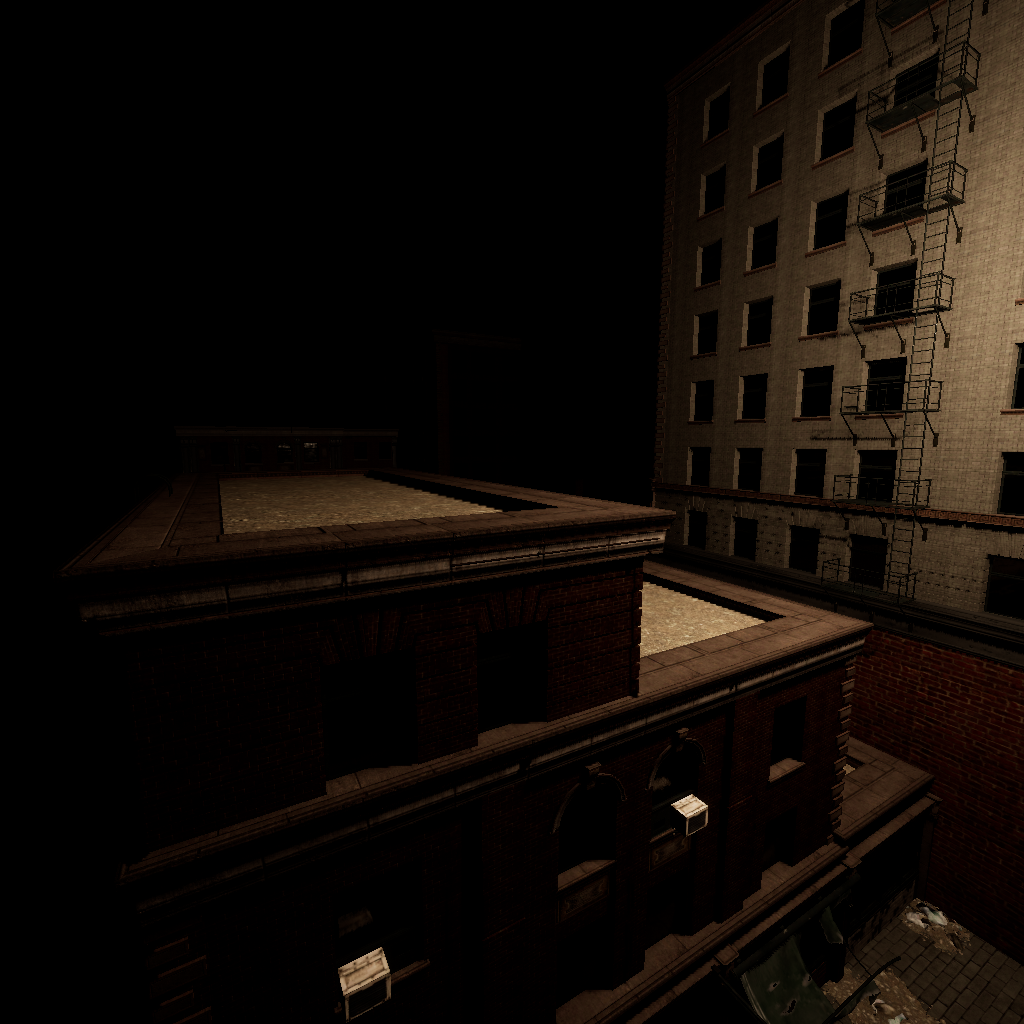}
    \end{tabular}
    \end{center}
    \vspace{-0.1cm}
    \caption{Visual results on the VIDIT dataset. Top: just 4500K, bottom: all temperatures.}
    \label{vidit_comparison}
    \vspace{-0.1cm}
    \end{figure}

The results of the experiments are presented in Table \ref{vidit} and Fig.\ \ref{vidit_comparison}. The supervised solution of Gafton and Maraz outperforms our model and the baseline in terms of PSNR. This is likely due to their supervised light direction classifier which is trained on VIDIT to provide the GAN with information regarding the input light direction. Apart from this addition, their generator follows the original pix2pix design. SILT also focuses on the target illumination but does not attempt to discern that of the source. This suggests that gaining more understanding of not only the reference but also the input illumination may lead to meaningful performance improvements. When it comes to other metrics, undisclosed by Gafton \& Maraz, we beat the baseline in terms of SSIM and the perceptual (VGG) score. 

The second set of experiments, trained on all temperatures and directions, tackles a more complex task as the networks now have to simultaneously deal with light colour and direction changes. Nevertheless, our solution outperforms the supervised model of Wang \etal\ in terms of SSIM and achieves a similar PSNR score. We also beat the baseline across all metrics. Visually, all methods struggle to reconstruct the dark region. Wang \etal\ and the pix2pix baseline attempt to lighten the area yet their reconstructions are very blurry and, while lighter, not similar to the ground truth. Our self-supervised SILT method manages to translate the colour setting but does not sufficiently lighten up the dark region. 
    \vspace{-0.2cm}
    \begin{table}[h]
    \begin{center}
    \begin{tabular}{ |l|c|c|c|  }
    \hline
    Model &  SSIM $\uparrow$ & PSNR $\uparrow$ & VGG $\downarrow$ \\
    \hline \hline 
    Gafton and Maraz \cite{gafton2020}&  n/a & \textbf{21.45} & n/a\\
    pix2pix &\underline{0.619} & 17.80 & \underline{0.350} \\
    SILT (ours)& \textbf{0.654} & \underline{17.88} & \textbf{0.341} \\
    \hline 
    Wang \etal \cite{wang_deep_2020}&   \underline{0.596} & \textbf{17.59} & n/a \\
    pix2pix & 0.447& 15.01 & \underline{0.419}\\
    SILT (ours)& \textbf{0.606}&\underline{17.00}&\textbf{0.371}\\
    \hline
    \end{tabular}
    \end{center}
    \caption{Results on the VIDIT dataset. Some metrics were not provided by the authors and are marked as not available (n/a).}
    \label{vidit}
    \vspace{-0.1cm}
    \end{table}
    
\vspace{-0.3cm}
\subsection{Cross-dataset results}
To verify the generality of our method, we test the performance of SILT on data from outside of its training dataset. For these experiments we choose the model trained on the Multi-Illumination dataset as we believe it represents real-life scenarios better than the dark VIDIT data. We show a few output images re-styled using SILT in Fig.\ \ref{generality}. For examples showing results obtained by running SILT on higher definition data, please see Appendix B. 

The images used for demonstrations in Fig.\ \ref{generality} come from VIDIT and Places \cite{zhou_2017_places} datasets. Since the style reference for the Multi-Illumination dataset is quite light and warm-toned, we can see such changes in the re-styled outputs. In VIDIT image a) more texture details are recovered, in comparison with the VIDIT-trained model (see: Fig.\ \ref{vidit_comparison}, top row). The model is similarly successful in re-styling the dark parking image from Places a). In the remaining images we can primarily see changes in the white balance.

    \begin{figure}[t]
    \begin{center}
    \begin{tabular}{c @{\hspace{0.05cm}} c @{\hspace{0.15cm}}c@{\hspace{0.05cm}}c@{\hspace{0.05cm}}c}
    \footnotesize VIDIT a) &\footnotesize VIDIT b) &\footnotesize Places a) &\footnotesize Places b) & \footnotesize Places c) \\
    \includegraphics[width=2.45cm]{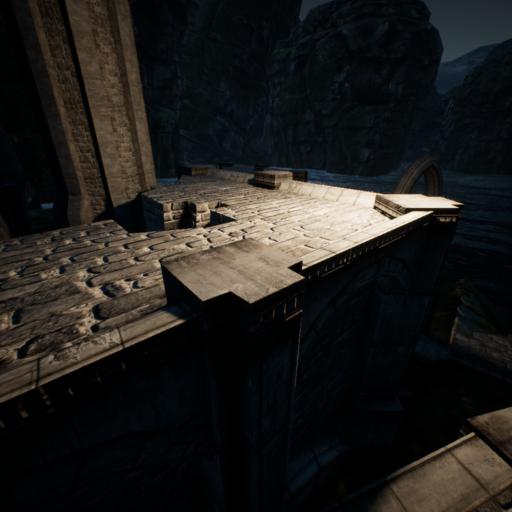}&
    \includegraphics[width=2.45cm]{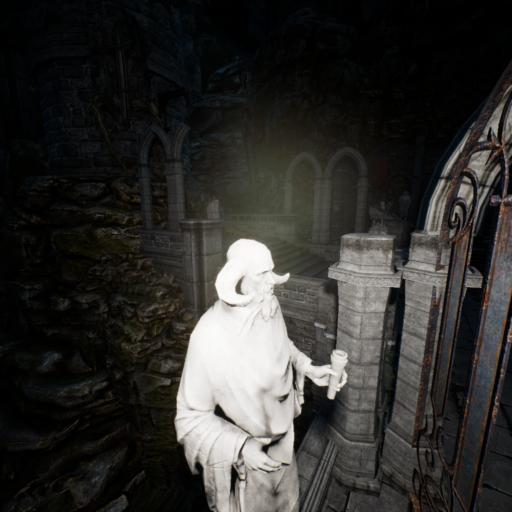}&
    \includegraphics[width=2.45cm]{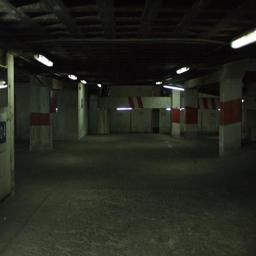}&
    \includegraphics[width=2.45cm]{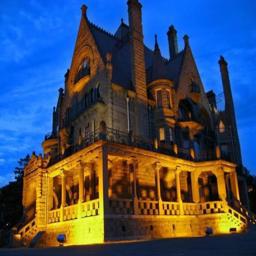}&
    \includegraphics[width=2.45cm]{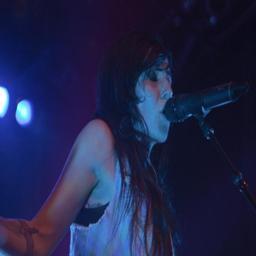}\\
    \includegraphics[width=2.45cm]{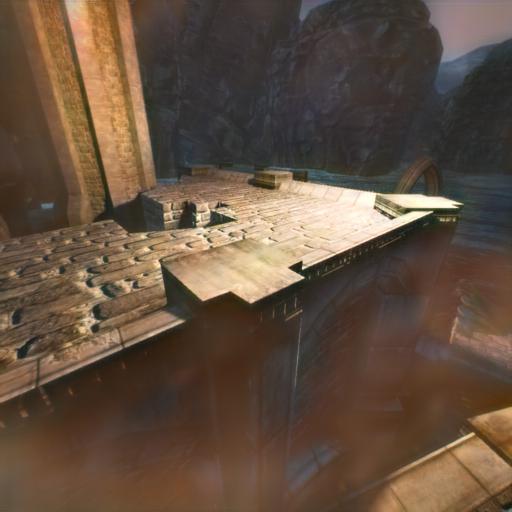}&
    \includegraphics[width=2.45cm]{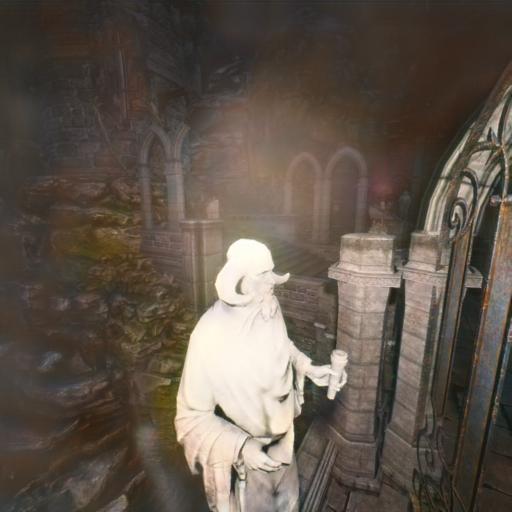}&
    \includegraphics[width=2.45cm]{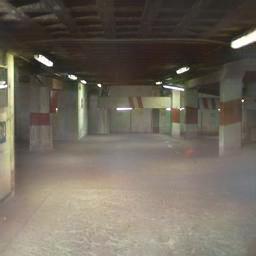}&
    \includegraphics[width=2.45cm]{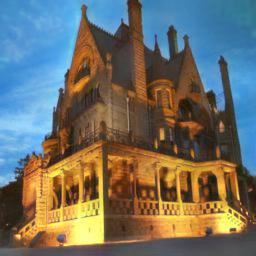}&
    \includegraphics[width=2.45cm]{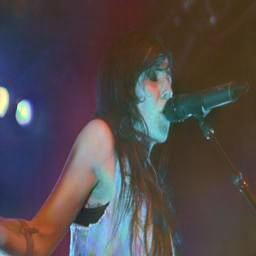}
    \end{tabular}
    \end{center}
    \caption{Visual results across different data using SILT trained on the Multi-Illumination dataset. The top row shows input images and the bottom row - their corresponding outputs.}
    \label{generality}
    \vspace{-0.3cm}
    \end{figure}

\section{Conclusions}
In this paper we proposed SILT, a self-supervised solution for general-purpose lighting transfer, a previously uncharted domain. We also demonstrated that using self-supervised image decomposition can lead to better physical accuracy in the re-styled images. To reap these gains, one can approach this process implicitly, and let the network learn what features to extract to optimise the downstream lighting style transfer task.

Unsurprisingly, our solution has some limitations. When we train the model on VIDIT and do not show it ground truth images (due to our self-supervised approach), SILT struggles to fill in the dark regions and instead only adjusts the colour/white balance (see Fig.\ \ref{vidit_comparison}). While the supervised methods are better at inserting textured content in such areas, their solutions are rarely fully accurate, and low-light region enhancement remains an open problem. 

Our future goals are to improve the lighting transfer accuracy and to expand the discussed solution to include people and/or more objects in our scenes. While advances in domain-specific relighting are being made, no existing approaches can cover real-life scenarios where full-bodied humans exist in complex environments. Therefore, we wish to fill this niche and move a step closer towards comprehensive lighting enhancement.
\newline\newline \textbf{\textcolor{bmv@sectioncolor}{Acknowledgements. }} This work was partially supported by the British Broadcasting Corporation (BBC) and the Engineering and Physical Sciences Research Council's (EPSRC) industrial CASE project ``Computational lighting in video'' (voucher number 19000034).

\newpage
\bibliography{main}
\newpage

\section*{Appendix A: Implicit image decomposition}
\begin{figure}[h]
    \begin{center}
    \begin{tabular}{c @{\hspace{0.05cm}} c @{\hspace{0.05cm}}c@{\hspace{0.05cm}}c}
    \footnotesize Input image & \footnotesize Reflectance map $R$&\footnotesize Shading map $S$&\footnotesize New shading map $\hat{S}$\\
    \includegraphics[width=3cm]{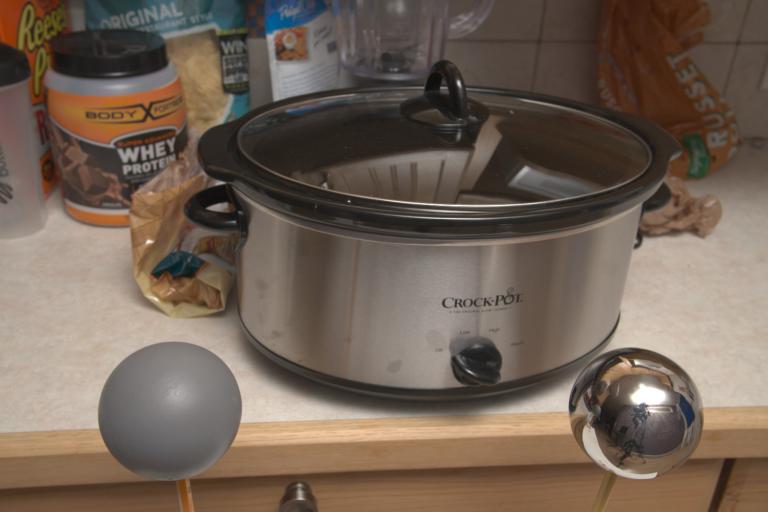}&
    \includegraphics[width=3cm]{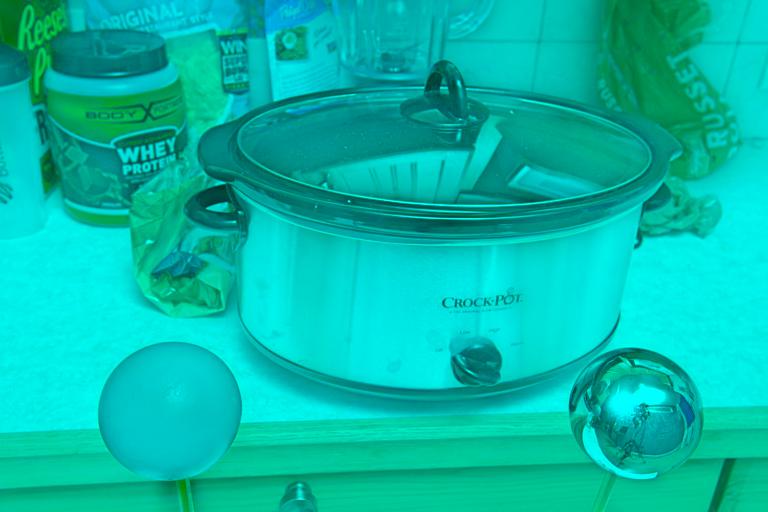}&
    \includegraphics[width=3cm]{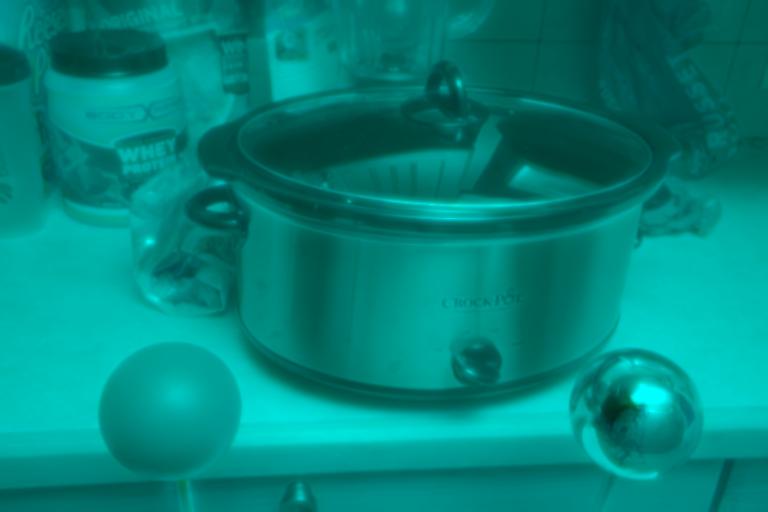}&
    \includegraphics[width=3cm]{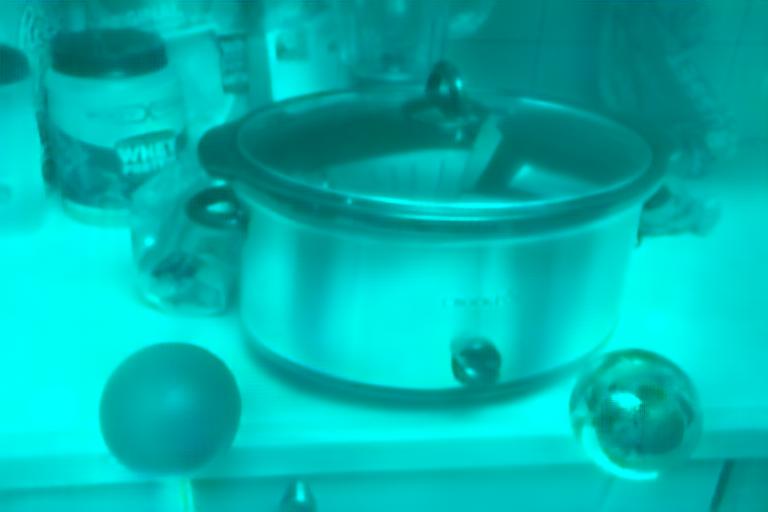}\\
    \includegraphics[width=3cm]{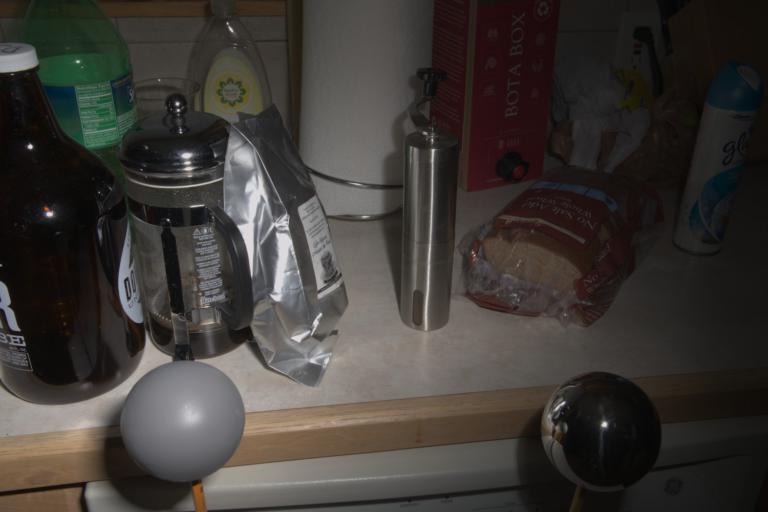}&
    \includegraphics[width=3cm]{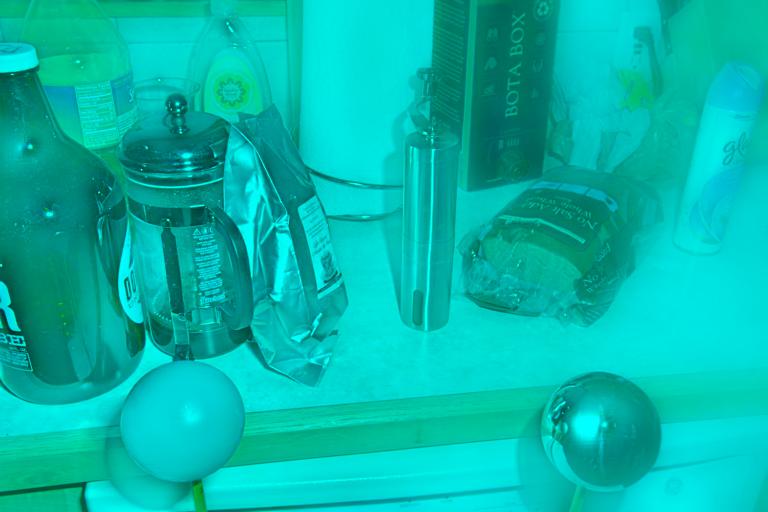}&
    \includegraphics[width=3cm]{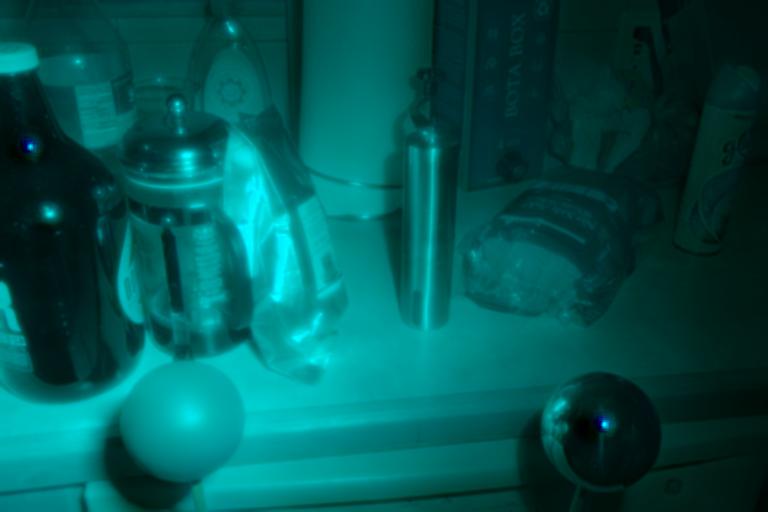}&
    \includegraphics[width=3cm]{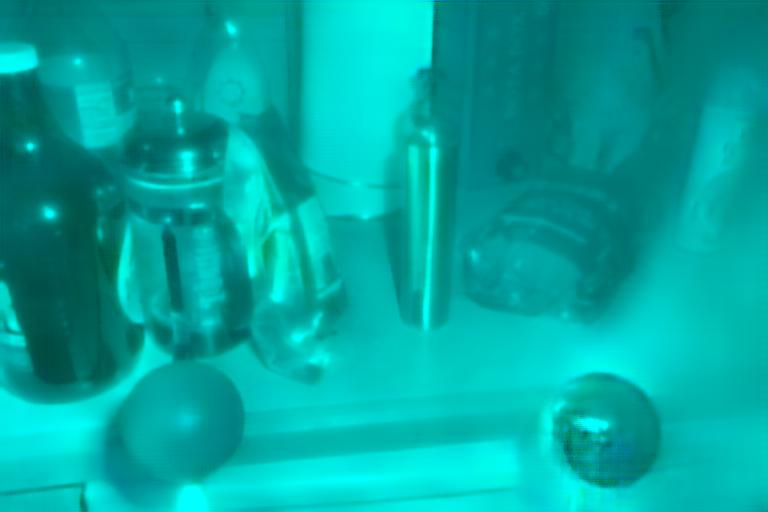}\\
    \includegraphics[width=3cm]{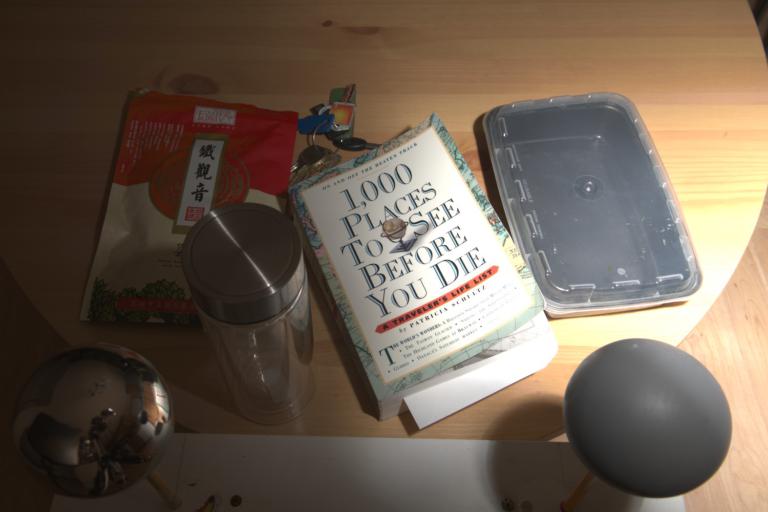}&
    \includegraphics[width=3cm]{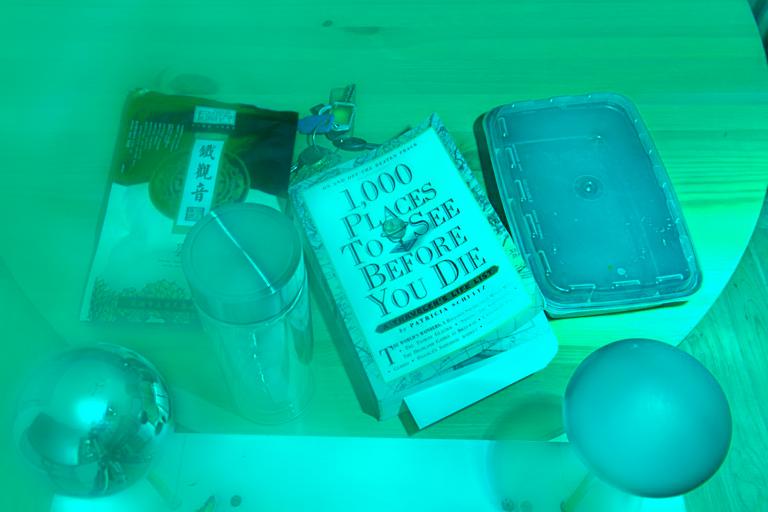}&
    \includegraphics[width=3cm]{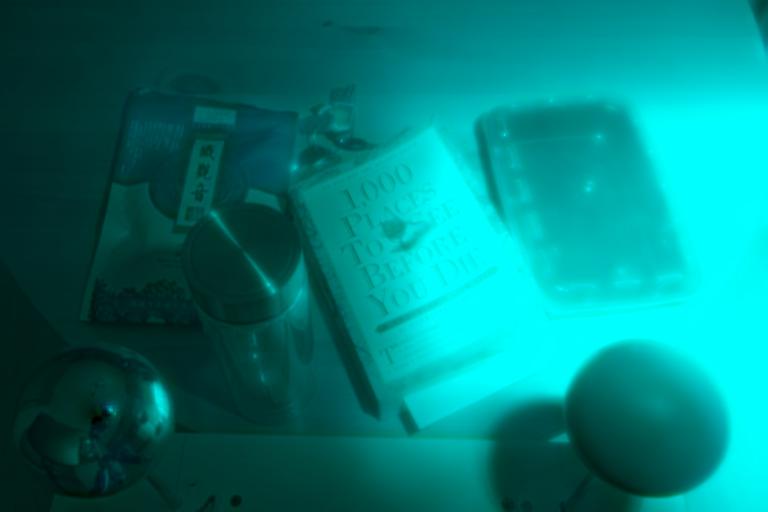}&
    \includegraphics[width=3cm]{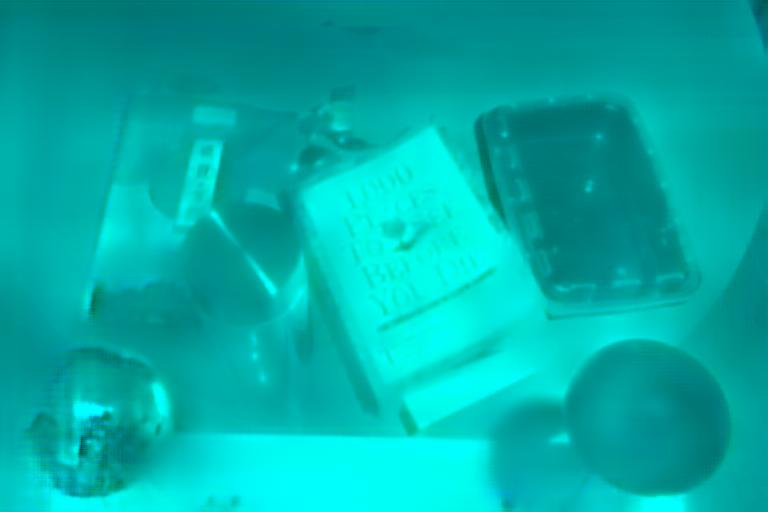}\\
    \end{tabular}
    \end{center}
    \caption{Examples of inputs, their implicitly decomposed reflectance and shading maps, and the new shading maps. Each row depicts a different scene and lighting direction.}
    \label{r_s_maps}
    \vspace{-0.5cm}
    \end{figure}

In this section we discuss the benefits of implicit image decomposition. We also show a few examples of reflectance $R$ and shading $S$ maps generated by $G_{\lambda}$ during the implicit decomposition step, as well as the new shading maps $\hat{S}$ acquired by processing the original shading map by generator $G$. The images are shown in Fig.\ \ref{r_s_maps}.

As can be seen in the examples in Fig.\ \ref{r_s_maps}, the new shading maps processed by generator $G$ do present a unified shading representation that differs from the original shading maps which correspond to different input light directions. This aligns with our goal of translating all images, regardless of initial lighting style, to the same domain. Additionally, we can observe clear shadows in the input shading maps, particularly distinct in the bottom two rows. This shows that the varying factors are recognised as such and correctly assigned to the shading (`changing-factors') maps.

We note that the reflectance and shading images have a visible cyan tint, but, as demonstrated by all examples shown in the main paper body, this is not reflected in the final, lighting-corrected images. We believe this is due to the white balance of the indoor lighting and the multiplicative colour space. Regardless, we wish to reiterate that achieving top performance in this domain was never our objective. We only care about performing decomposition in a way that simplifies the lighting transfer task for the generator $G$ and, thus, improves the overall performance of our model (as shown in the ablation study). 
    
\section*{Appendix B: Higher definition examples}
\begin{figure}[h]
    \begin{center}
    \begin{tabular}{c @{\hspace{0.05cm}} c}
    \footnotesize Input 3600$\times$2400 &\footnotesize Output 3600$\times$2400 \\
    \includegraphics[width=6cm]{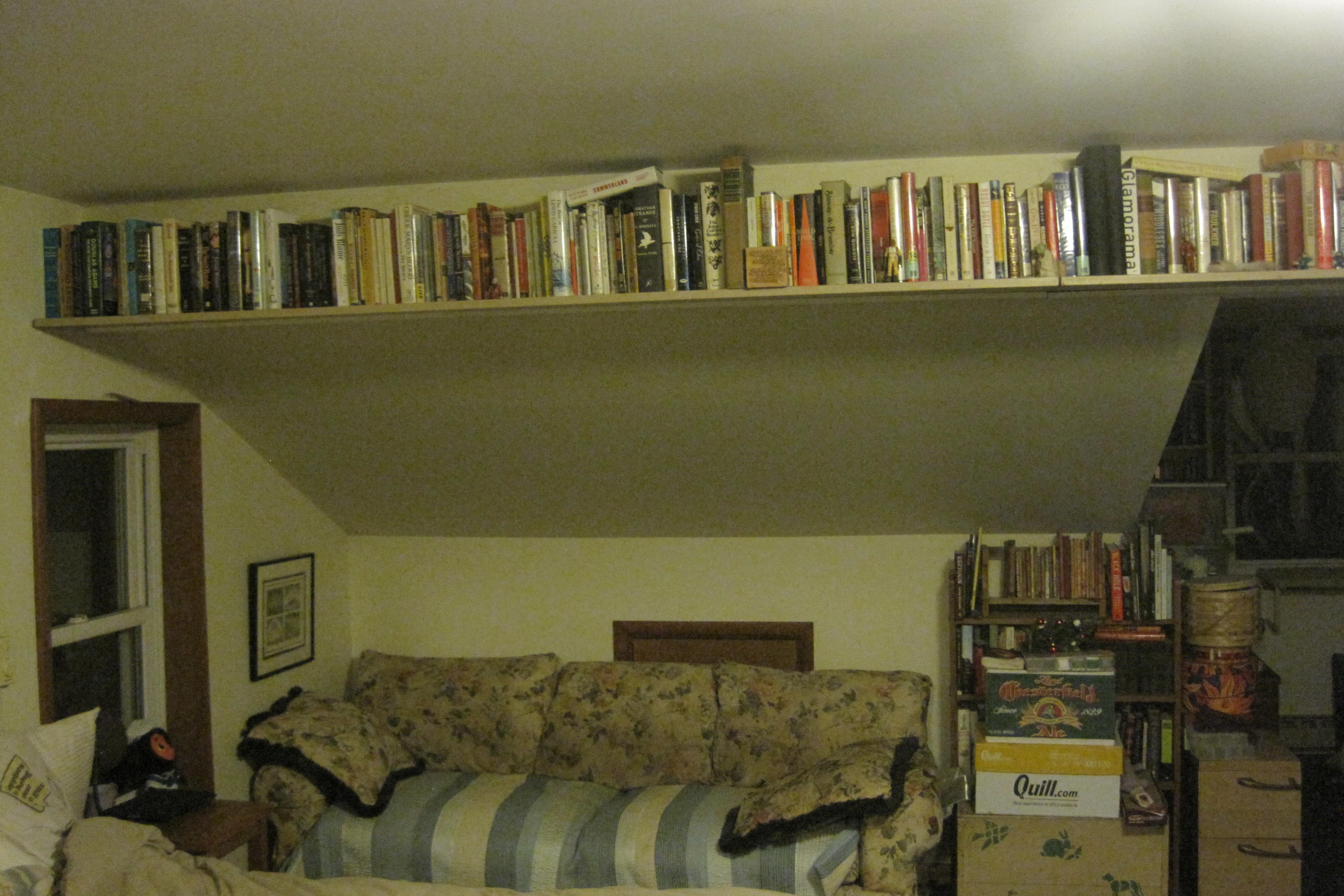}&
    \includegraphics[width=6cm]{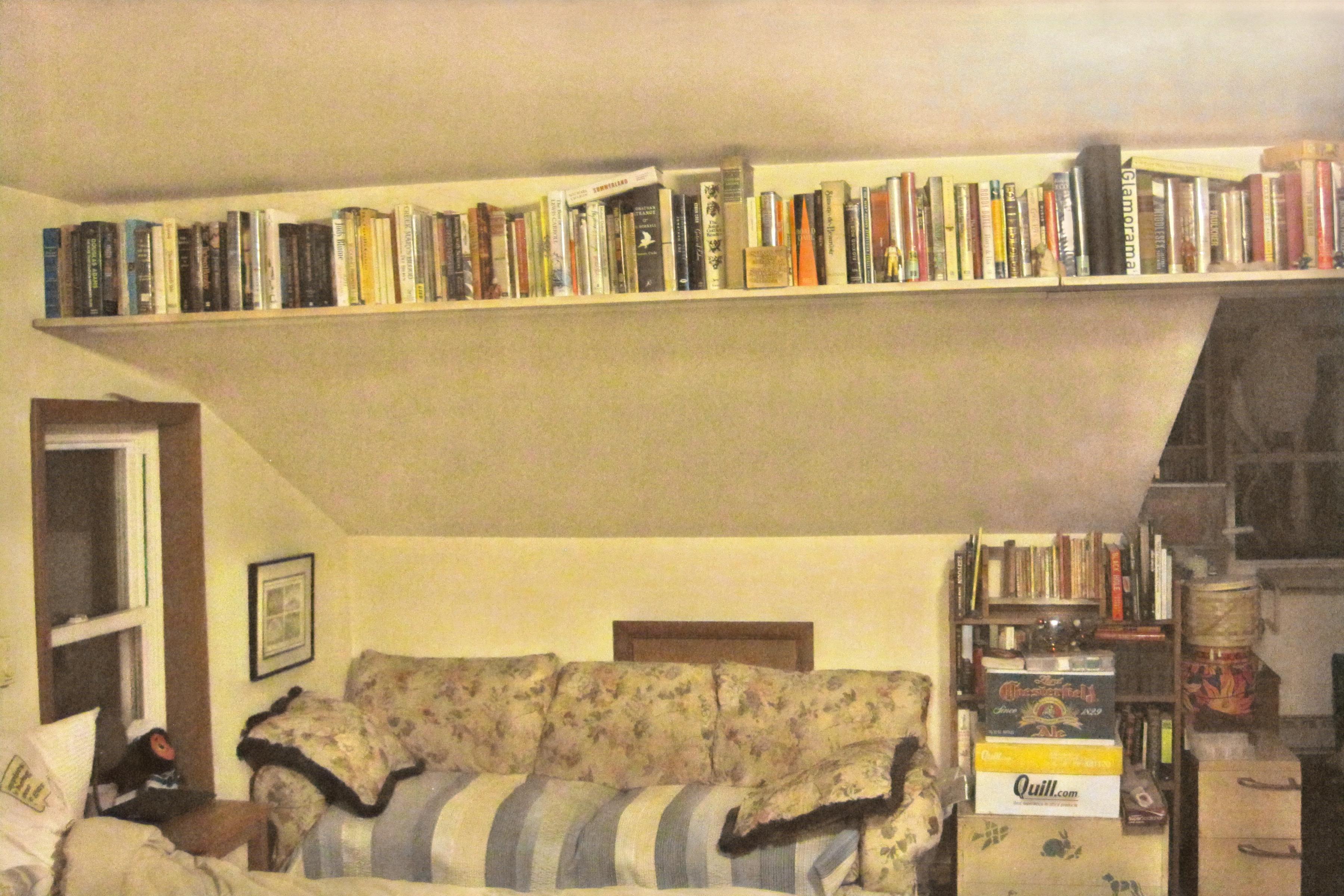}\\
    \footnotesize Output 2400$\times$1600 &\footnotesize Output 1800$\times$1200 \\
    \includegraphics[width=6cm]{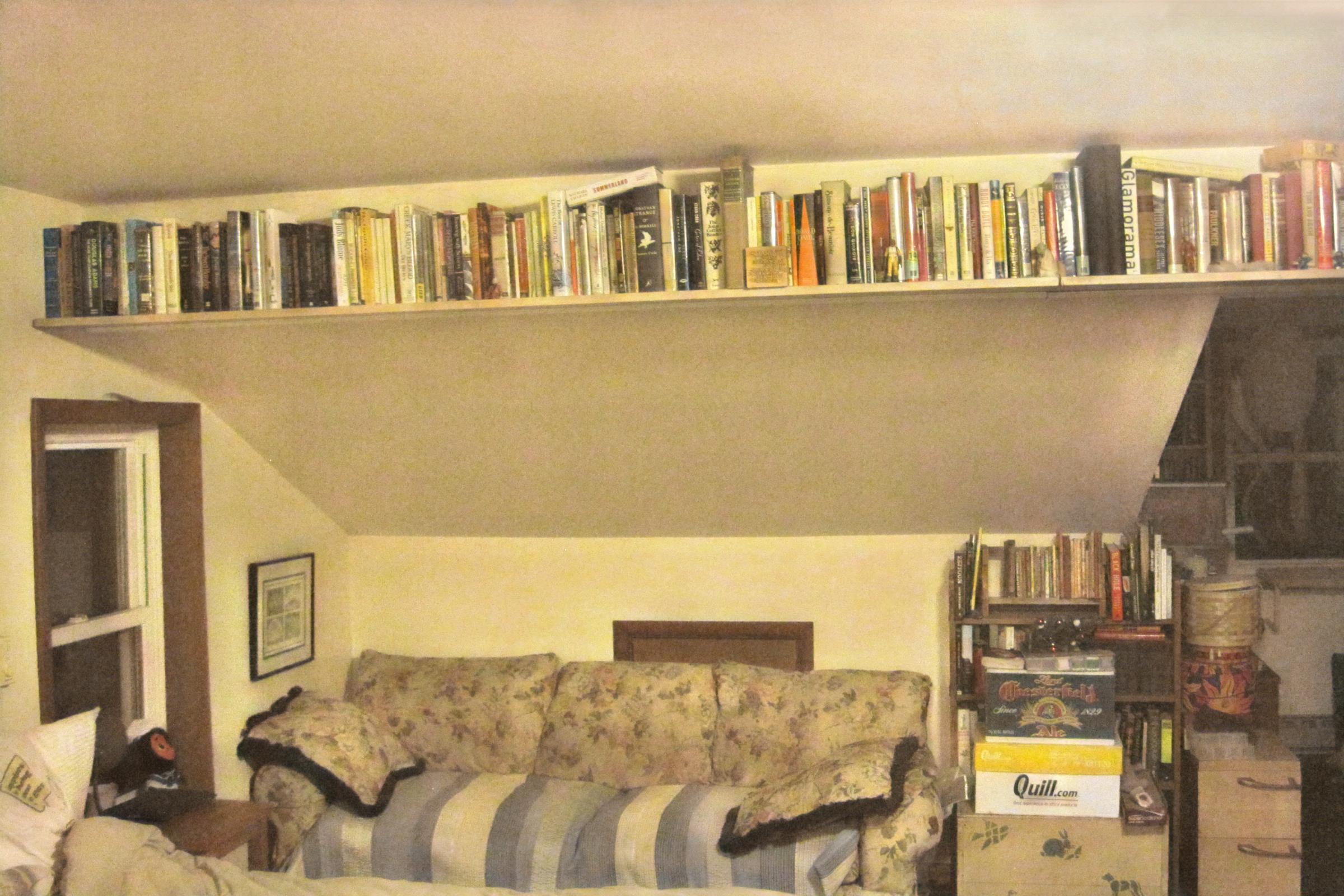}&
    \includegraphics[width=6cm]{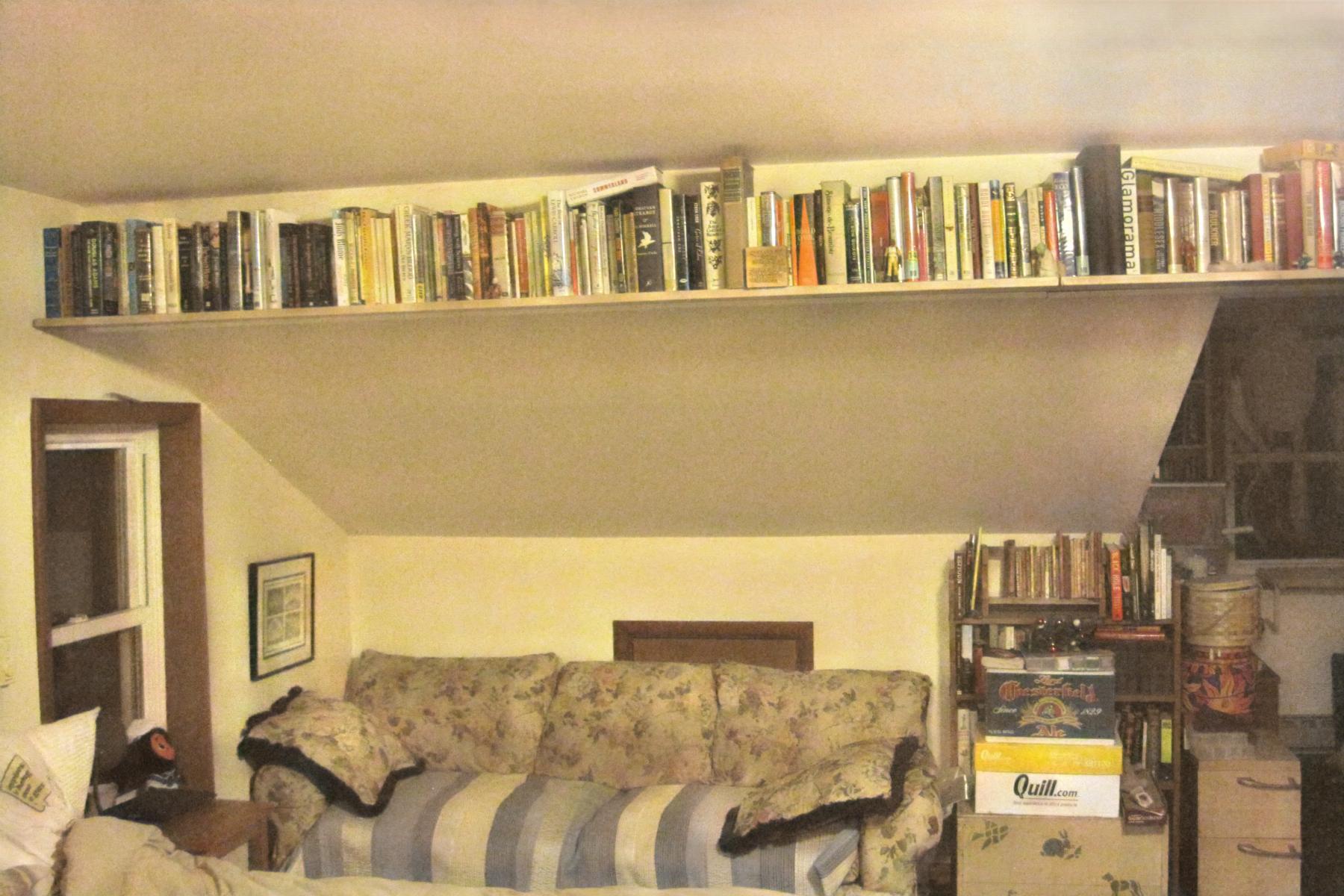}\\
    \footnotesize Output 1200$\times$800 &\footnotesize Output 600$\times$400 \\
    \includegraphics[width=6cm]{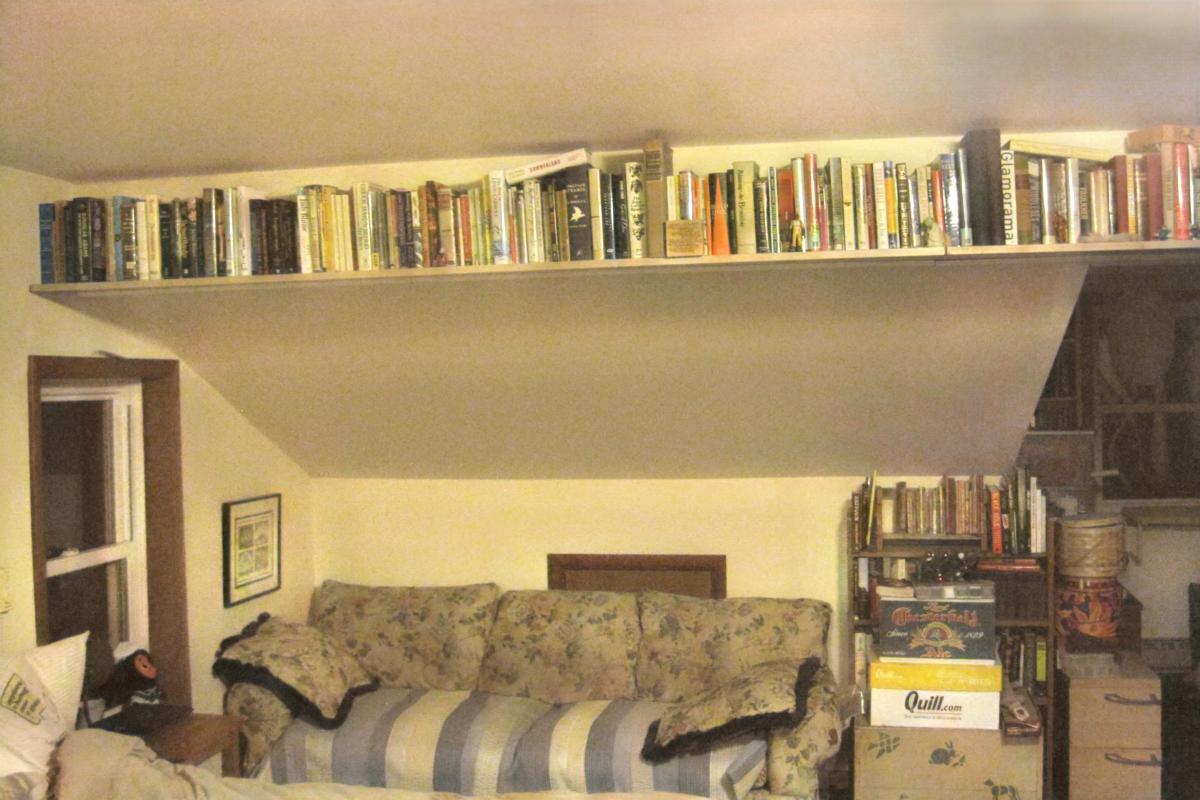}&
    \includegraphics[width=6cm]{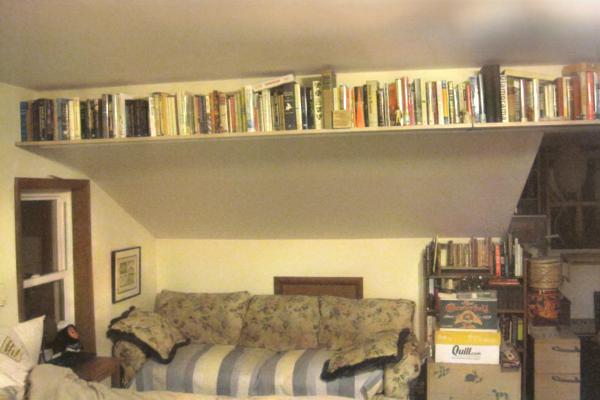}\\
    \end{tabular}
    \end{center}
    \caption{Lighting transfer results for an OpenSurfaces image re-styled using SILT trained on the Multi Illumination dataset. The numbers next to `Output' show the size of the input image used to create the output image (of the same size).}
    \label{HD}
    \end{figure}

\begin{figure}[h]
    \begin{center}
    \begin{tabular}{c @{\hspace{0.05cm}} c}
    \footnotesize Input 2048$\times$1024 &\footnotesize Output 2048$\times$1024 \\
    \includegraphics[width=6cm]{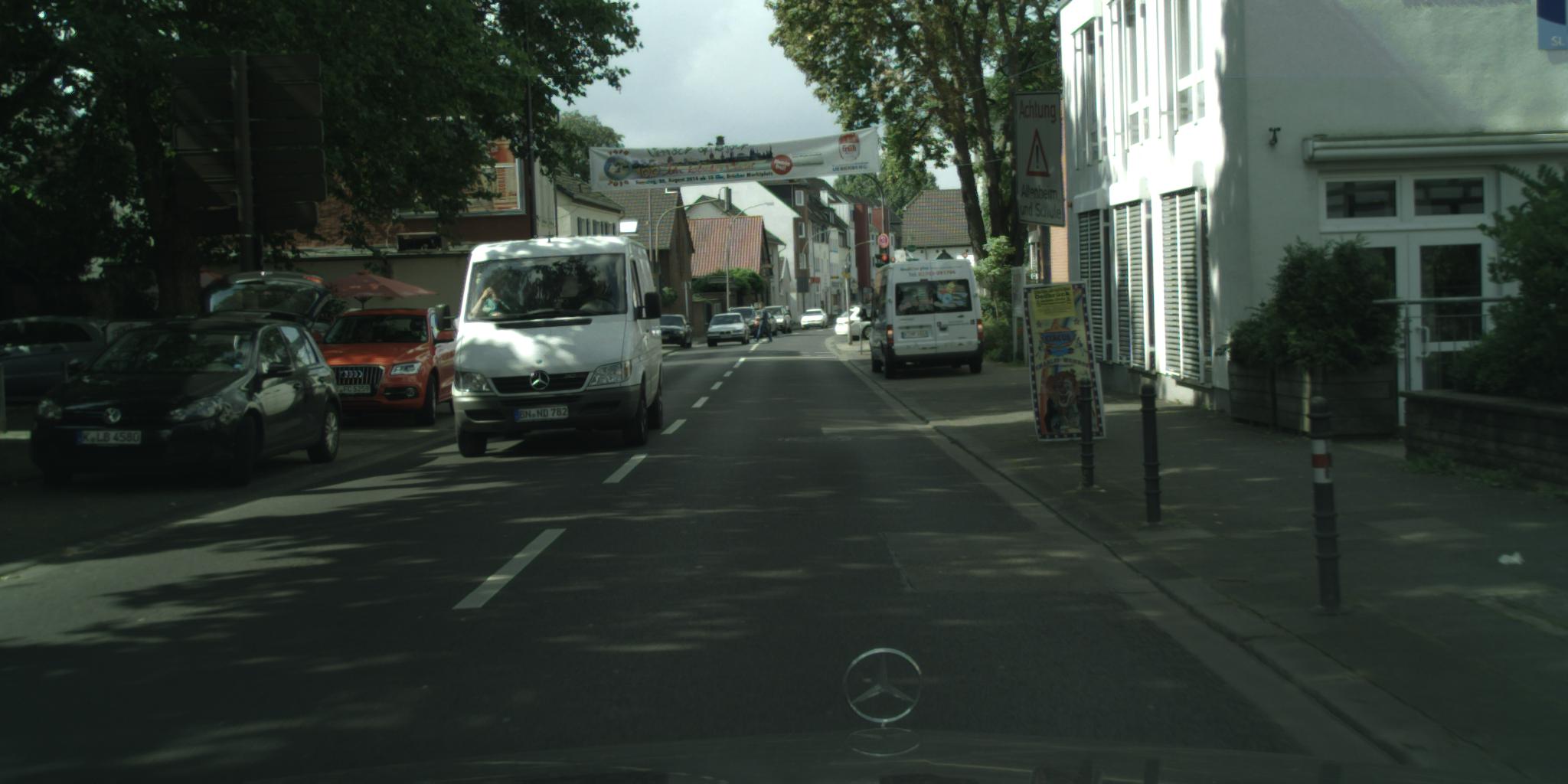}&
    \includegraphics[width=6cm]{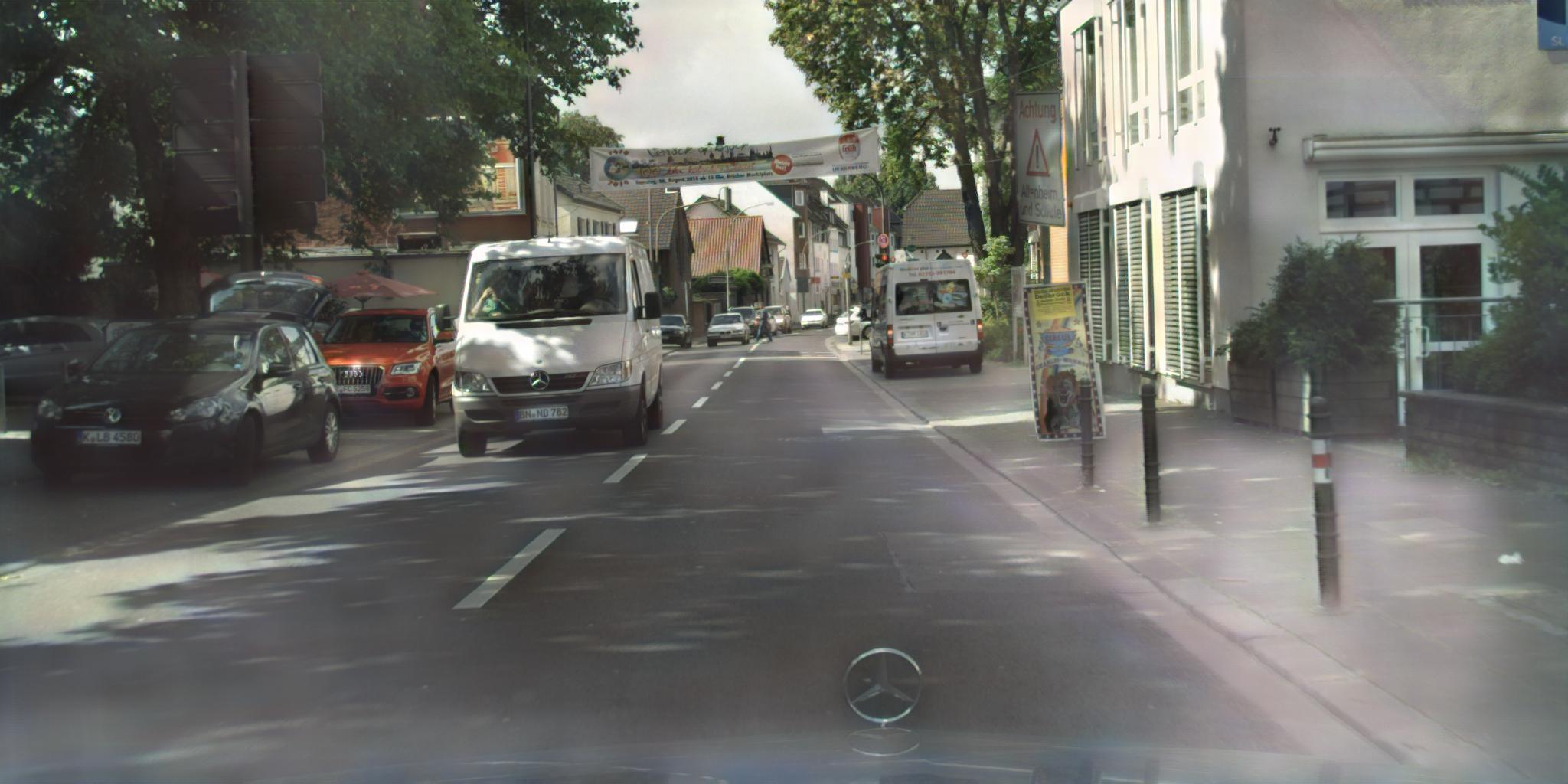}\\
    \footnotesize Output 1024$\times$512 &\footnotesize Output 512$\times$256 \\
    \includegraphics[width=6cm]{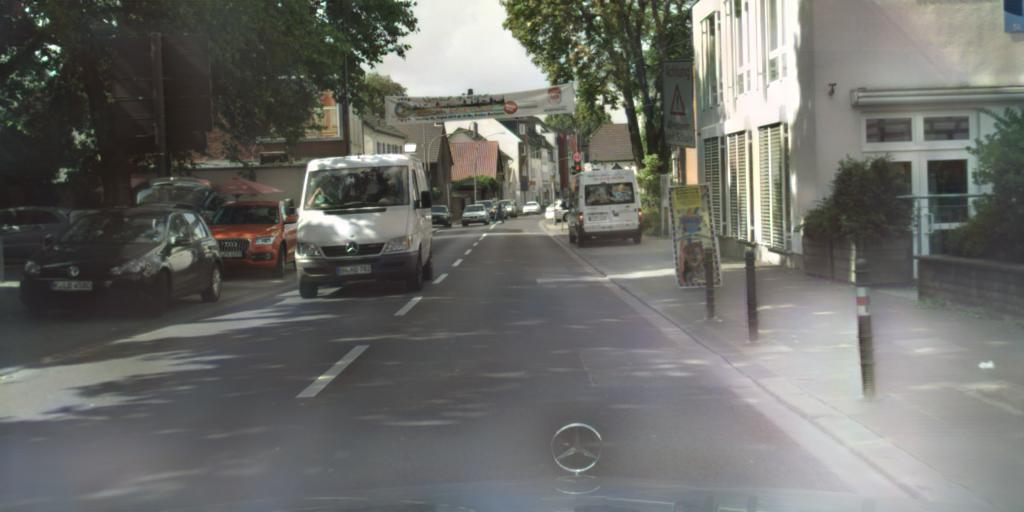}&
    \includegraphics[width=6cm]{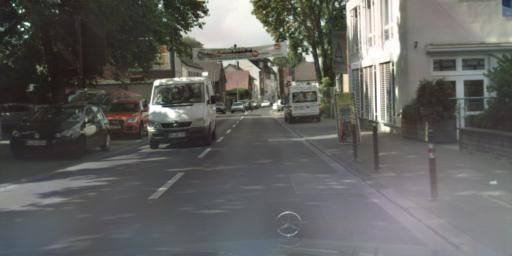}\\
    \end{tabular}
    \end{center}
    \caption{Lighting transfer results for a Cityscapes image re-styled using SILT trained on the Multi Illumination dataset. The numbers next to `Output' show the size of the input image used to create the output image (of the same size).}
    \label{HD2}
    \end{figure}

The max.\ size of images used for training models discussed in the main body of the paper was 786$\times$512 pixels. In case of the Multi Illumination \cite{murmann19} data, the produced outputs were additionally cropped to highlight the most interesting image areas. In this section, however, we show that SILT (trained on the Multi Illumination dataset) can produce visually pleasing results across different image sizes. We demonstrate this by applying SILT to a number of higher quality images and then performing a sweep across different image resolutions. The samples are presented in Fig.\ \ref{HD} and \ref{HD2} and come from the OpenSurfaces \cite{bell13opensurfaces} and Cityscapes \cite{cityscapes} datasets,  respectively. 

Even though SILT was trained on lower-definition images, the model can apply lighting changes to larger images without serious artefacts. In the largest re-styled images, the titles of larger books (OpenSurfaces) and vehicle registration plates (Cityscapes) are still as clearly visible as in the input image. In Fig.\ \ref{HD}, the two smallest images have the most visible artefacts, particularly in the top right corner. In Fig.\ \ref{HD2}, however, this effect appears to be worse for the two larger images, where a road pole on the right hand side is surrounded by white `haze'. \looseness -1

The pix2pixHD paper \cite{pix2pixHD} recommends using a LocalEnhancer addition to the regular generator, to allow for HD image processing. In case of 4k data, two of such enhancers would need to be trained and finetuned with the best model. However, we find that even without these add-ons our model does not visibly degrade the input image quality and performs well regardless of input size.

\section*{Appendix C: Model complexity}
We report the complexity of our SILT model, specifically - the number of total vs trainable parameters, and GFLOPs. These are measured for both datasets and image sizes used for training SILT, and shown below in Table \ref{performance}. 


    \begin{table}[h]
    \begin{center}
    \begin{tabular}{ |c|c|c|  }
    \hline
    dataset + image size&  GFLOPs & \# model params (of which trainable)\\
    \hline \hline
    Multi Illumination \cite{murmann19} (786$\times$512) &  2792.768 & \multirow{2}{*}{79,328,552 (66,383,592)}   \\
    VIDIT \cite{vidit} (512$\times$512)&1862.846&  \\
    \hline
    \end{tabular}
    \end{center}
    \caption{SILT model complexity for Multi Illumination and VIDIT datasets and their corresponding training image sizes.}
    \label{performance}
    \vspace{-0.3cm}
    \end{table}
    

\section*{Appendix D: Output similarity loss}
In the main paper body we report the results obtained using the output similarity loss $\mathcal{L}_{os}$. The loss is defined in Eq.\ \ref{os_loss} and composed of the standard L1 loss and the L1 loss between spatial gradients calculated over the generated images (referred to as L1(sg) in Table \ref{output_similarity}). 

In Table \ref{output_similarity} we show that adding L1(sg) to the commonly used L1 error metric improves the model performance. We start with just the core losses (\textit{i.e.} GAN losses $\mathcal{L}_g$ and $\mathcal{L}_d$ and the decomposition loss $\mathcal{L}_{dcp}$). When we add the L1 loss, we can observe a significant decrease across all performance metrics. When L1 is combined with L1(sg), the decline is smaller and the PSNR value actually improves slightly. 

    \begin{table}[h]
    \begin{center}
    \begin{tabular}{ |c|c|c|c|c|  }
    \hline
    \# & losses used&  SSIM $\uparrow$& PSNR $\uparrow$&VGG $\downarrow$\\
    \hline \hline
    1& core &0.730	&17.075	&0.428 \\
    2&core + L1 &0.632&15.352& 0.547 \\
    3&core + L1 + L1(sg) &0.682 &17.142 & 0.495 \\
    \hline
    \end{tabular}
    \end{center}
    \caption{$\mathcal{L}_{os}$: The effect of adding the L1(spatial gradient) sub-loss to the L1 error metric. }
    \label{output_similarity}
    \vspace{-0.2cm}
    \end{table}

\end{document}